\documentclass[conference]{IEEEtran}
\usepackage{times}

\usepackage[numbers]{natbib}
\usepackage{multicol}
\usepackage[bookmarks=true]{hyperref}
\usepackage{graphicx}
\usepackage{booktabs} 
\usepackage{longtable}
\usepackage{caption}
\usepackage{adjustbox}
\usepackage{subcaption}
\usepackage{csquotes}
\usepackage{amsmath}
\usepackage{xtab}
\usepackage{makecell}
\usepackage{csquotes}

\begin{document}

\title {Neural Scaling Laws in Robotics}


\author{
\authorblockN{Sebastian Sartor}
\authorblockA{TUM\\
Munich, Germany\\
Email: sebastian.sartor@tum.de}
\and
\authorblockN{Neil C. Thompson}
\authorblockA{MIT\\
Cambridge, MA, US\\
Email: neil\_t@mit.edu}
}


%

\maketitle

\begin{abstract}
Neural scaling laws have driven significant advancements in machine learning, particularly in domains like language modeling and computer vision. However, the exploration of neural scaling laws within robotics has remained relatively underexplored, despite the growing adoption of foundation models in this field. This paper represents the first comprehensive study to quantify neural scaling laws for Robot Foundation Models (RFMs) and Large Language Models (LLMs) in robotics tasks. Through a meta-analysis of 327 research papers, we investigate how data size, model size, and compute resources influence downstream performance across a diverse set of robotic tasks. Consistent with previous scaling law research, our results reveal that the performance of robotic models improves with increased resources, following a power-law relationship. Promisingly, the improvement in robotic task performance scales notably faster than language tasks. This suggests that, while performance on downstream robotic tasks today is often moderate-to-poor, increased data and compute are likely to signficantly improve performance in the future.  Also consistent with previous scaling law research, we also observe the emergence of new robot capabilities as models scale.
\end{abstract}

\IEEEpeerreviewmaketitle

\section{Introduction}

Significant advancements in deep learning over recent years have been primarily driven by scaling—training larger neural networks on increasing amounts of data with ever-growing computational resources~\citep{alabdulmohsin2022revisiting}. The concept that scaling neural networks leads to improved performance dates back to research conducted as early as 2010~\citep{coates2011analysis}. However, it was not until around 2017 that researchers like Jonathan Rosenfeld~\citep{rosenfeld2019constructive, rosenfeld2019relation, rosenfeld2021predictability, rosenfeld2021scaling}, teams at Baidu~\citep{hestness2017deep}, and later OpenAI~\citep{henighan2020scaling, kaplan2020scaling} formally articulated “neural scaling laws.” These laws describe how the performance (e.g. the loss of neural networks) systematically improves as a function of model size, data size, and compute resources. The concept of neural scaling laws aligns with the principles articulated in the Bitter Lesson of 2019, which highlights the critical role of scalable computation in achieving superior performance~\citep{sutton2019bitter}.


Neural scaling laws have been studied extensively in various domains, ranging from language modeling to vision and reinforcement learning, following a power-law function (e.g.,~\citet{kaplan2020scaling, hoffmann2022training, zhai2022scaling, hilton2023scaling}). They not only provide a framework for understanding how neural network architectures and data distributions impact performance, but have also proven to be beneficial for planning sample sizes, particularly in data-scarce domains, and have helped reduce the environmental footprint of experimentation~\citep{sorscher2022beyond, muennighoff2024scaling}. Overall, scaling laws facilitate the identification of optimal scaling coefficients, allow performance prediction based on given inputs, and enable estimating the required inputs for desired performance outcomes~\citep{alabdulmohsin2022revisiting}.

Inspired by the capabilities and high generalizability of foundation models in domains like language and vision, robotics researchers are exploring their application to the physical world, envisioning general-purpose robotics and a potential resolution to Moravec's Paradox (e.g.,~\citet{brohan2023rt, reed2022generalist, padalkar2023open}).

Traditionally, roboticists developed specialized models tailored to specific applications, robots, and environments, often using multi-module system architectures. Recently, the field has shifted toward training large, generalist, pre-trained policies end-to-end. These Vision-Language-Action (VLA) models are designed for efficient adaptation across diverse robots, tasks, and environments~\citep{padalkar2023open, EricJang}. 

Another significant trend is integrating foundation models trained on internet-scale data—such as Vision-Language Models (VLMs) and Large Language Models (LLMs) into robotic tasks. This approach enhances robots' ability to interpret natural language commands and visually understand tasks~\citep{padalkar2023open}. By bridging high-level reasoning with low-level control, these models enable robots to generalize across tasks and environments, demonstrating semantic reasoning capabilities such as understanding and executing commands like \enquote{stack the red block on top of the green cylinder.}

Despite these advances, neural scaling laws in robotics remain mainly unexplored. Previous research hints that scaling principles may hold, but no comprehensive quantification has been conducted (e.g.,~\citet{padalkar2023open, brohan2023rt}). Another weakness of previous scaling law work is that it focuses on measures that do not directly translate into real-world task performance. This is particularly important because identifying and quantifying scaling laws in the context of robotics provides a crucial framework for developing general-purpose robotic systems. Scaling laws enable researchers to predict performance outcomes, allocate resources more efficiently, and ensure adaptability across tasks. By understanding these principles, we can streamline experimentation, reduce costs, and enhance the environmental sustainability of robotics research. 

This study addresses this gap by identifying neural scaling laws for robotics. Specifically, we study \textbf{1) Do scaling laws observed in other domains like language and vision hold true for RFM and LLMs in robotics in terms of data size, model size and compute resources? 2) What are the characteristic power-law coefficients? and 3) Do RFM and LLMs in robotics exhibit emergent capabilities similar to those observed in other domains?}


\section{Background and related work}

\begin{figure}
    \centering
    \begin{subfigure}[b]{0.49\textwidth}
        \includegraphics[width=\textwidth]{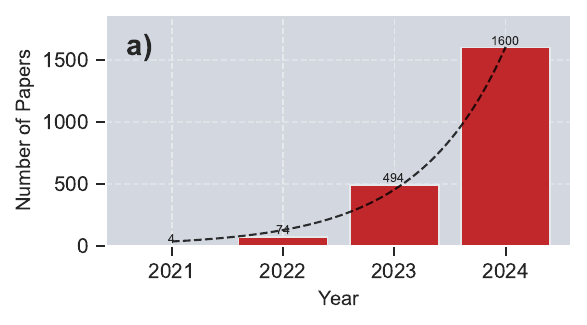}
        \label{fig:year}
    \end{subfigure}
    \hfill
    \begin{subfigure}[b]{0.49\textwidth}
        \includegraphics[width=\textwidth]{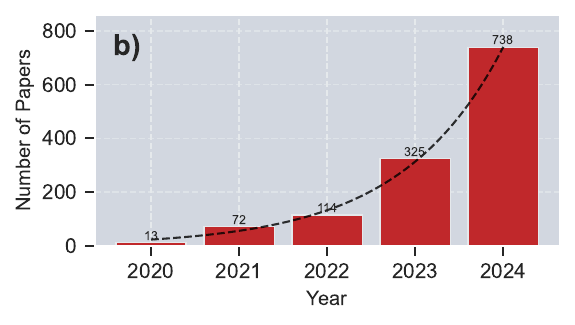}
        \label{fig:scale_year}
    \end{subfigure}
    \caption{Growth of Robotics (a) and Scaling Laws (b) research over time}
    \label{fig:research_growth}
\end{figure}

\paragraph{Robotics \& Embodied AI} The field of robotics has struggled for some time with challenges related to generalizability and scalability, as developing machine intelligence for autonomous systems in the physical world is both costly and time-intensive. Existing solutions are typically designed for specific tasks, limiting their ability to generalize across a wide array of applications~\citep{survey_grid}. Building on the remarkable success of foundation models in domains like language and vision, researchers and entrepreneurs have begun exploring how to extend these advancements to the field of robotics. These models can be divided into two primary categories~\citep{hu2023toward}. The first category includes Robot Foundation Models (RFMs), which are large, versatile models trained on diverse datasets and designed to generalize across various downstream robotic tasks. These robotics models often incorporate various learning strategies, such as imitation learning, behavior cloning, reinforcement learning, and diffusion policies. Many of them, especially those that handle large-scale data or multimodal inputs, are built upon architectures like the Transformer \citep{vaswani2017attention}. RFMs are being applied in robotics across various domains, including low- and high-level perception, planning, and data augmentation~\citep{Kawaharazuka_2024}. Drawing inspiration from the broader concept of foundation models in AI~\citep{bommasani2021opportunities}, RFMs can be further classified into three types according to~\citet{Kawaharazuka_2024}: 1) 
Pre-trained Visual Representations (PVRs): Models such as R3M~\citep{nair2022r3m} and RPT~\citep{radosavovic2023robot} focus on learning visual features that can be applied to multiple robotic tasks; 2) Vision-Language Models (VLMs): These models, like Palm-E~\citep{driess2023palm}, integrate visual and textual information to enhance robotic reasoning and understanding; and 3) Vision-Language-Action (VLA) Models: Also known as Robot Transformers or Large Behavior Models (LBMs), those models are designed for end-to-end learning of control policies and dynamics (e.g., RT-2, $\pi_0$, and Open-VLA)~\citep{black2024pi_0, kim2024openvla, brohan2023rt}.  The second approach besides RFMs integrates LLMs into robotics, leveraging their semantic reasoning to bridge the gap between language and physical action~\citep{zeng2023large, li2024foundation}. Overall, research on foundation models for robotics is a relatively recent research field and has demonstrated exponential growth in the past, as illustrated in Figure~\ref{fig:research_growth}\footnote{Citation data is based on Google Scholar, using the keywords “robot learning” and “foundation models” (Dec. 2024)}. Beginning with just 4 papers in 2021, the number of publications surged to 2,172 by 2024, with 74\% of these papers published in 2024 alone. This trend highlights the rapidly growing interest and progress in the domain of robotics research with foundation models.\\
While these strategies hold great promise, they are hindered by several significant challenges. The lack of internet-scale robotics datasets remains a major obstacle, as the high cost and complexity of data collection coupled with the vast diversity of robot types, environmental conditions, and task requirements make high-quality data generation an expensive and time-consuming process. While the volume of data is important, its diversity is equally crucial to ensure robust and adaptable model training~\citep{etukuru2024robot}. Oftentimes data is collected through manual teleoperation in the real world which scales linearly with human time ($< 24 \, \text{hrs/robot/day}$). Therefore, another active research field is centered around world foundation models and large-scale robotic simulations such as Nvidia Cosmos, RoboCasa, or Genesis~\citep{genesis, robocasa2024}. Their data generation scales with the amount of compute, yet faces the hurdle of the \enquote{sim-to-real gap}, where models trained in simulated environments struggle to perform effectively in the unpredictable and complex conditions of the real world, highlighting the intricate challenges posed by real-world robot-environment interactions. Besides data generation, evaluation and benchmarking are challenging given the cost, time and diversity of conditions to test the system's capabilities~\citep{zhou2023train, survey_firoozi2023foundation}. Lastly, important concepts from AI such as in-context learning and prompt engineering are still to be figured out for robotics.\\

\paragraph{Neural Scaling Laws} Neural scaling laws are empirical principles that describe the relationship between a model's performance and factors such as its model size, the size of its training dataset, and its compute resources. These laws typically follow a power-law function, often with a cross-entropy objective, indicating that as model size, data size and compute increase, the model's performance improves predictably. The general form of the power law described in~\citep{kaplan2020scaling} for data size is given by:

\[
L(D) = \left( \frac{A}{D} \right)^{\alpha},
\]

for model size by:

\[
L(N) = \left( \frac{B}{N} \right)^{\beta}, \text{ and}
\]

for compute by:

\[
L(C) = \left( \frac{F}{C} \right)^{\gamma},
\]

where \(L(N)\), \(L(D)\), and \(L(C)\) represent the loss for model, data size, and compute, respectively, \(N\) is the model size, \(D\) is the data size, \(C\) is the compute, and \(A\), \(B\), and \(F\) are constants. The exponents \(\alpha\), \(\beta\), and \(\gamma\) describe the scaling behavior with respect to model size, data size, and compute. The choice of performance metric depends on the task and modality, ranging from negative log-likelihood per token and perplexity to accuracy or mean squared error. Among the parameters in the scaling law, \(\alpha\), \(\beta\), and \(\gamma\) play central roles.
It quantifies the efficiency of scaling, determining how rapidly performance improves as resources increase. Higher values of \(\alpha\), \(\beta\), and \(\gamma\) indicate more efficient scaling, where larger improvements are achieved with less additional investment in size, data, or compute. Tables~\ref{tab:scaling_laws_kaplan} and~\ref{tab:scaling_laws_henighan} provide an overview of typical \(\alpha\), \(\beta\), and \(\gamma\) values observed across various modalities. Interestingly, the language domain stands out as significantly less efficient compared to other domains~\citep{kaplan2020scaling, henighan2020scaling}.

\begin{table}[t]
\caption{Power law exponent of neural scaling laws for language models, as detailed in~\citet{kaplan2020scaling}}
\label{tab:scaling_laws_kaplan}
\centering
\begin{tabular}{lccc}
\toprule
\textbf{Domain} & \textbf{Compute} & \textbf{Data} & \textbf{Model Size} \\
\midrule
Language & -0.050 & -0.076 & -0.095 \\
\bottomrule
\end{tabular}
\end{table}

\begin{table}[!t]
\caption{Compute and Model Size scaling laws in other domains, as detailed in~\citet{henighan2020scaling}}
\label{tab:scaling_laws_henighan}
\centering
\begin{tabular}{p{1.8cm}cc}
\toprule
\textbf{Domain} & \textbf{Model Size} & \textbf{Compute} \\
\midrule
\makecell{Language} & \(\frac{N}{1.47 \times 10^{14}}^{-0.070}\) & \(\frac{C}{3.47 \times 10^8}^{-0.048}\) \\
\makecell{Image 8x8} & \(3.12 + \frac{N}{8.0 \times 10^1}^{-0.24}\) & \(3.13 + \frac{C}{1.8 \times 10^{-8}}^{-0.19}\) \\
\makecell{Image 32x32} & \(2.20 + \frac{N}{6.3 \times 10^1}^{-0.13}\) & \(2.21 + \frac{C}{3.6 \times 10^{-9}}^{-0.10}\) \\
\makecell{Image \\ VQ 32x32} & \(3.07 + \frac{N}{1.9 \times 10^4}^{-0.14}\) & \(3.17 + \frac{C}{2.6 \times 10^{-6}}^{-0.12}\) \\
\makecell{Text-to-Im \\(Text)} & \(\frac{N}{5.6 \times 10^8}^{-0.037}\) & \((\text{combined text/image loss})\) \\
\makecell{Text-to-Im \\(Image)} & \(2.0 + \frac{N}{5.1 \times 10^3}^{-0.16}\) & \(1.93 + \frac{C}{1.5 \times 10^{-6}}^{-0.15}\) \\
\makecell{Im-to-Text \\ (Text)} & \(\frac{N}{7.0 \times 10^8}^{-0.039}\) & \((\text{combined text/image loss})\) \\
\makecell{Im-to-Text \\(Image)} & \(2.0 + \frac{N}{5.5 \times 10^3}^{-0.15}\) & \(1.97 + \frac{C}{1.5 \times 10^{-6}}^{-0.15}\) \\
\makecell{Video \\VQ 16x16} & \(1.01 + \frac{N}{3.7 \times 10^4}^{-0.24}\) & \(0.95 + \frac{C}{2.2 \times 10^{-5}}^{-0.14}\) \\
\makecell{Math \\(Extrapolate)} & \(0.28 + \frac{N}{1.1 \times 10^4}^{-0.16}\) & \(0.14 + \frac{C}{1.4 \times 10^{-5}}^{-0.17}\) \\
\bottomrule
\end{tabular}
\end{table}

Notably, more scaling is not always better. For example, the Chinchilla language model demonstrated that optimal performance could be achieved with a more balanced allocation of data and compute resources, reducing inefficiencies in scaling~\citep{hoffmann2022training}. They introduce a parametric fit that accounts for both model size and data size, given by the following expression:

\[
L (D, N) = \left( \frac{A}{D} \right)^{\alpha_D} + \left( \frac{B}{N} \right)^{\beta_N} + E,
\]
where \(E\) is a constant.

Scaling laws have been extensively studied across various domains in machine learning, including LLMs, image and video generative modeling, and reinforcement learning~\citep{kaplan2020scaling, henighan2020scaling, hilton2023scaling}. Most research has focused on pretraining and fine-tuning phases, but recently, test-time compute (or inference scaling) has also emerged as an active area of investigation \citep{snell2024scalingllmtesttimecompute}. As models scale, both in size and in the volume of data they are trained on, they exhibit not only quantitative improvements but also the emergence of novel qualitative behaviors, often referred to as emergent capabilities~\citep{wei2022emergent, schaeffer2024emergent}. These emergent abilities are absent in smaller models or those trained on limited datasets, arising in a way that cannot be directly predicted from the performance of smaller configurations.

While scaling laws provide high predictability in upstream performance improvements, the downstream capabilities of scaled models often remain challenging to predict~\citep{ganguli2022predictability}. 

The field of scaling laws is growing rapidly, as illustrated in Figure~\ref{fig:research_growth}\footnote{The citation count presented here is based on data from Google Scholar, which includes publications that cite \citet{kaplan2020scaling} and contain the keyword \enquote{Scaling Law} (Dec. 2024)}. Since the release of the seminal paper on neural scaling laws by~\citet{kaplan2020scaling} in 2020, research in this area has gained significant momentum. Despite substantial progress, it remains an intensely researched area with numerous open questions, both in breadth and depth.

\paragraph{Neural Scaling Laws in robotics}
While scaling laws have been extensively studied in domains such as language models and vision models, their exploration in robotics remains largely uncharted territory. Although some recent work suggests that scaling phenomena hold for robotics regarding data size (e.g., \citep{brohan2022rt, padalkar2023open, brohan2023rt}), model size (e.g., \citep{bousmalis2023robocat, majumdar2024we}), and compute (e.g., \citep{liu2023tail}), where larger models tend to perform better, very few studies explicitly mention \enquote{scaling laws} or systematically quantify scaling behavior, as shown in Table~\ref{tab:scaling_robot}. We identified only four papers that mention scaling laws or examine them in the context of robotics and embodied AI. First, \citet{villasevil2024scaling} discusses scaling laws in terms of increasing the number of environments in a dataset. However, they neither quantify these laws nor define a specific functional form. Their study focuses on two tasks (pick-and-place) using a dataset with a maximum of 56 scenes. Second, \citet{lin2024data} investigates data scaling laws for single-task robot policies and their generalization. They emphasize the importance of data diversity over the sheer number of demonstrations and find that performance approximately follows a power law with diminishing returns. The power-law exponent varies between -0.446 and -0.844 for the number of objects and environments. Third, \citet{duan2024aha} examines the performance of a VLM with respect to dataset size. Their analysis reveals an average quadratic fit gradient of 0.0022 across four metrics, indicating a scaling effect. This suggests that further increasing the dataset size could enhance model performance. Finally, \citet{pearce2024scaling} explores scaling laws related to agents and world models, focusing on upstream compute scaling for behavior cloning and world models. They observe power-law relationships with exponents ranging from -0.03 to -0.31. Their findings highlight how task, architecture, and tokenizer choices heavily influence these coefficients. Additionally, they propose optimal dataset and model sizes in relation to compute resources. Overall, while these studies provide early insights, further research is needed to address the significant gaps in understanding scaling laws within robotics and embodied AI.

\begin{table}[ht]
\caption{Summary of Scaling Studies in robotics \& Embodied AI}
\label{tab:scaling_robot}
\centering
\small
\renewcommand{\arraystretch}{1.5} 
\begin{tabular}{lcp{2.7cm}p{2.5cm}}
\hline
\textbf{Study} & \textbf{Date} & \makecell{\textbf{Focus}} & \makecell{\textbf{Perf. (Range)}} \\
\hline
\makecell{\citep{villasevil2024scaling}} & \makecell{June '24} & \makecell{Robotics\\Data (env.)} & \makecell{Mentioned it}\\[4pt]
\makecell{\citep{lin2024data}} & \makecell{Oct '24} & \makecell{Robotics\\Data (obj./env.)} & \makecell{Downstream;\\Power Law\\(-0.446, -0.844)\\} \\[12pt]
\makecell{\citep{duan2024aha}} & \makecell{Oct '24} & \makecell{Robotics\\Data (general)} & \makecell{Downstream;\\Quadratic\\} \\[8pt]
\makecell{\citep{pearce2024scaling}} & 
\makecell{Nov '24} & 
\makecell{Agents/ \\World Models\\Compute} & 
\makecell{Upstream;\\Power Law\\(-0.03, -0.31)} \\
\bottomrule
\end{tabular}
\end{table}

In contrast to the prior studies, our work expands the understanding of scaling laws in robotics and embodied AI by addressing several key gaps. We analyze a significantly larger dataset, enabling robust insights that cover the full spectrum of scaling behaviors. Our findings confirm that power-law approximations effectively model the observed trends across diverse scenarios. Unlike previous studies, we examine scaling laws comprehensively across data, model size, and compute. Additionally, we contextualize these laws by comparing them with those observed in other modalities. Our analysis differentiates between model types, such as LLMs and VLAs, deployment contexts (simulation versus real-world robotics), and even variations in compute scaling, including pretraining versus fine-tuning. Furthermore, we explore task- and method-specific effects, shedding light on nuanced scaling dynamics. Finally, we identify the emergence of novel capabilities that manifest as models scale, providing critical insights into the transformative potential of these systems.



\section{Empirical approach}
\subsection{Research paper meta-analysis}
This study presents a meta-analysis of 327 research papers, drawing from diverse sources such as survey articles~\citep{survey_firoozi2023foundation, survey_xiao2023robot, survey_grid}, GitHub repositories~\citep{GitHub_robotics-fm-survey, GitHub_Awesome-LLM-Robotics, GitHub_Everything-LLMs-And-Robotics, GitHub_Awesome-Robotics-Foundation-Models}, citations of the seminal work by~\citep{padalkar2023open}, and recent publications featured in newsletters and personalized news feeds (e.g., X and~\citep{MLPaperofTheWeek}). The analyzed papers encompass a wide range of RFMs and LLMs applied in robotics, addressing tasks such as manipulation, navigation, reasoning, planning, and instruction following. Studies focusing solely on navigation are excluded, unless combined with manipulation tasks due to the defined focus of this study. These models vary in scope, from task-specific systems to versatile general-purpose architectures, and are implemented across diverse robotic embodiments, including industrial arms and legged robots, operating in both simulated and real-world environments. A notable focus is on manipulation tasks, primarily involving single-arm operations such as grasping and pick-and-place actions, which are prevalent in the largest robotics datasets, such as RT-X, DROID or AgiBot World~\citep{padalkar2023open, khazatsky2024droid, AgibotWorld}. These tasks are typically set in everyday environments, like kitchens and tabletops, though some studies extend to specialized domains, such as laboratory automation.

\begin{figure}
    \centering
    \begin{subfigure}[b]{0.49\textwidth}
        \includegraphics[width=\textwidth]{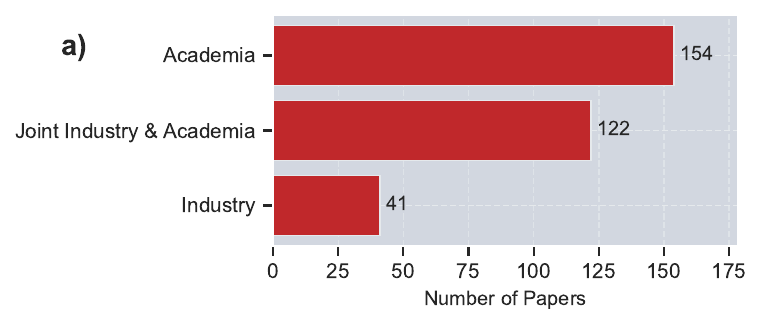}
        \label{fig:affiliation}
    \end{subfigure}
    \hfill
    \begin{subfigure}[b]{0.49\textwidth}
        \includegraphics[width=\textwidth]{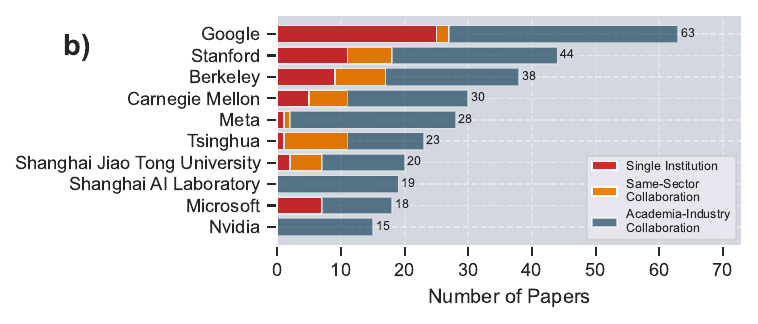}
        \label{fig:institutions}
    \end{subfigure}
    \caption{Publication by sector (a) and top 10 research institutions by publications.}
    \label{fig:sector}    
\end{figure}

76\% of all 327 analyzed papers were published in 2023 and 2024, aligning with the exponental surge in publications in the field as illustrated in Figure~\ref{fig:research_growth}. Moreover, a significant portion of these studies originates from collaborative projects between industry and academia (Figure~\ref{fig:sector}), highlighting the increasing role of industry in AI research~\citep{ahmed2023growing}. Notably, Google DeepMind contributes to 19\% of all publications, making it the most prolific institution in this field, ahead of other leading entities like Stanford and UC Berkeley (Figure~\ref{fig:sector}).

We chose meta-analysis as our approach due to the nascent and exploratory nature of the robot learning field, which is still evolving toward standardization. This field is characterized by a diverse array of tasks, model architectures, and embodiment forms, making it challenging for individual experimental studies to provide a comprehensive understanding. Single studies often offer a narrow perspective and can be influenced by specific biases or constraints. By aggregating data across a wide variety of research efforts, meta-analysis allows us to uncover broader trends and insights that individual studies cannot capture.


\subsection{Data extraction from papers}
Each of the 327 papers was screened for relevance and evaluated for scalability studies across several metrics: Success Rate, computational resources (FLOP, PF-days), data usage (tokens, demonstrations, trajectories, episodes, images, frames), and model size (parameters). We focused exclusively on works that provided numeric performance values for the success rate, either directly in the paper or shared by the researchers in response to our inquiries, while acknowledging the existence of additional scaling studies using other metrics. The choice of success rate as the focus of our analysis is due to its status as the most commonly used metric.
We found that 13\% of the papers include some form of scalability study, as shown in table~\ref{tab:research_papers_per_metric}\footnote{The following papers provided the scaling law studies for this paper: \\
\textbf{Data:}~\citet{reed2022generalist, rana2024affordancecentricpolicylearningsample, wen2023any, shridhar2022cliport, gao2024efficientdatacollectionrobotic, tang2024embodimentagnosticactionplanningobjectpart, jiang2023vima, li2024llarasuperchargingrobotlearning, fu2024mobile, stone2023open, li2022pre, nair2022r3m, kuang2024ramretrievalbasedaffordancetransfer, brohan2022rt, radosavovic2023real, bousmalis2023robocat, radosavovic2023robot, etukuru2024robot, bu2024synergisticgeneralizedefficientdualsystem, zhou2023train}\

\textbf{Model size:}~\citet{staroverov2023fine, cheang2024gr2generativevideolanguageactionmodel, jiang2023vima, lynch2020language, huang2023instruct2act, stone2023open, chen2023open, wu2023tidybot, radosavovic2023real, radosavovic2023robot, wen2024tinyvlafastdataefficientvisionlanguageaction}\

\textbf{Compute:}~\citet{liu2023tail}
}. 

\begin{table}[ht]
\caption{Number of scaling studies found out of 327 papers in total}
\label{tab:research_papers_per_metric}
\centering
\begin{tabular}{lp{3cm}p{3cm}}
\toprule
\textbf{Metric} & \textbf{Papers with scale studies (with 3+ data points)} & \textbf{Scale studies extracted (with 3+ data points)} \\
\midrule
Data & 26 (20) & 275 (147) \\
Model Size & 22 (11) & 125 (43) \\
Compute & ~1 & ~~24 (24) \\
\bottomrule
\end{tabular}
\end{table}

These studies typically explore the effects of scaling data size, model size, and compute often through ablations comparing these factors to downstream performance metrics such as success rate. In many cases, a paper reports multiple scaling studies, such as those conducted for different tasks. Each scaling study is then fitted to a power law, meaning the number of power laws we analyze statistically corresponds to the number of scaling studies with at least three data points, the minimum required for modeling a power law with a constant. To clarify, we do not fit a single model to all scaling studies. Instead, we fit separate models for each scaling study presented in the papers. While conditions can vary between scaling studies, the conditions within each study are consistent. This approach allows us to derive multiple models, which we then compare statistically. It is also important to note that our study does not aim to estimate scaling laws for predicting model performance based on specific inputs. Instead, our focus is on understanding how models generally improve under the conditions presented in each study and evaluating the efficiency of these scaling studies. However, we found that many of these scaling studies are limited by their small number of data points, often consisting of only two or three data points.

The relatively small number of studies reporting scaling analyses underlines that this is still an early stage in the exploration of scalable solutions in robotics. This mirrors the progress seen in other machine learning domains, where scalable methods have played a significant role in advancing the field. Currently, most scaling studies focus on data (26 papers), followed by model size (22 papers) and compute (1 paper). Often, papers report multiple scalability analyses, assessing model performance across different tasks with varied inputs (see table~\ref{tab:research_papers_per_metric}). The scarcity of scalability studies related to computational resources underscores an important gap in the field. Previous research on LLMs indicates that model performance is not solely dependent on data, model size, or compute resources in isolation, but rather on the interplay among these factors~\citep{kaplan2020scaling}. Hence, in this paper we advocate for reporting computational resource usage~\citep{sevilla2023please}, as well as conducting comprehensive studies on computational scalability, potentially together with a compute-optimal scaling law study.

In robotics, there is no standard benchmark that ensures comparability of results across studies, and task success rate is the most commonly used performance metric. To analyze this metric effectively, we categorize the tasks as \enquote{seen} (familiar), \enquote{unseen} (unfamiliar), or as having unreported performance.
Most studies utilized demonstrations, trajectories, and episodes as their primary data sources. Given that these are the only metrics with substantial sample sizes, all data quantifications presented in this analysis are based exclusively on these samples. In contrast, there is a notable deficiency in computational studies; only one paper has explored how the success rate varies with changes in the number of epochs. Using the hardware specifications and training duration details provided in this paper, we estimated the training FLOP with a tool\footnote{\href{https://epochai.org/blog/estimating-training-compute}{https://epochai.org/blog/estimating-training-compute}} by Epoch.ai. FLOP, or PetaFLOP-days, serve as more precise metrics for comparing the computational demands of AI model training than, for example, GPU days~\citep{thompson2020computational, OpenAI_compute, Epoch_Compute, Heim_Compute}. 
\\

\subsection{Scaling Laws analysis} 
Using the dataset we compiled, we modeled the scaling behavior observed in the studies using the following power law equations:

\[
Error Rate(D) = \left( \frac{A}{D} \right)^{\alpha}+E,
\]

for data size (\(D\)),

\[
Error Rate(N) = \left( \frac{B}{N} \right)^{\beta}+E,
\]

for model size (\(N\)), and

\[
Error Rate(C) = \left( \frac{F}{C} \right)^{\gamma}+E,
\]

for compute (\(C\)).


Here, \(A\), \(B\), and \(F\) are scaling constants, while \(\alpha\), \(\beta\), and~\(\gamma\) are the scaling exponents corresponding to each resource. The error rate, expressed as a percentage (\(100 - \text{success rate}\)), decreases as resources increase. The additive constant \(E\) represents an offset value.

Among these coefficients, the scaling exponents (\(\alpha\),~\(\beta\),~and~\(\gamma\)) are particularly significant as they govern the rate at which performance improves with additional resources. Our analysis focuses on the regime where each of the scaling exponents falls within the range \(-1 < \alpha, \beta, \gamma < 0\), which accounts for 87\% of the scaling studies we reviewed. This range indicates diminishing but positive returns as resources grow, aligning with practical expectations.


\section{Results}
\subsection{Data Scaling Laws}
The scarcity of internet-scale robotics data remains a significant barrier to developing fully generalizable and high-performing robotics models. This limitation also explains why data scaling dominates the research landscape, accounting for 69\% of the 214 scaling studies with at least three data points reviewed.

\begin{figure}[!t]
    \centering
    \includegraphics[width=0.5\textwidth]{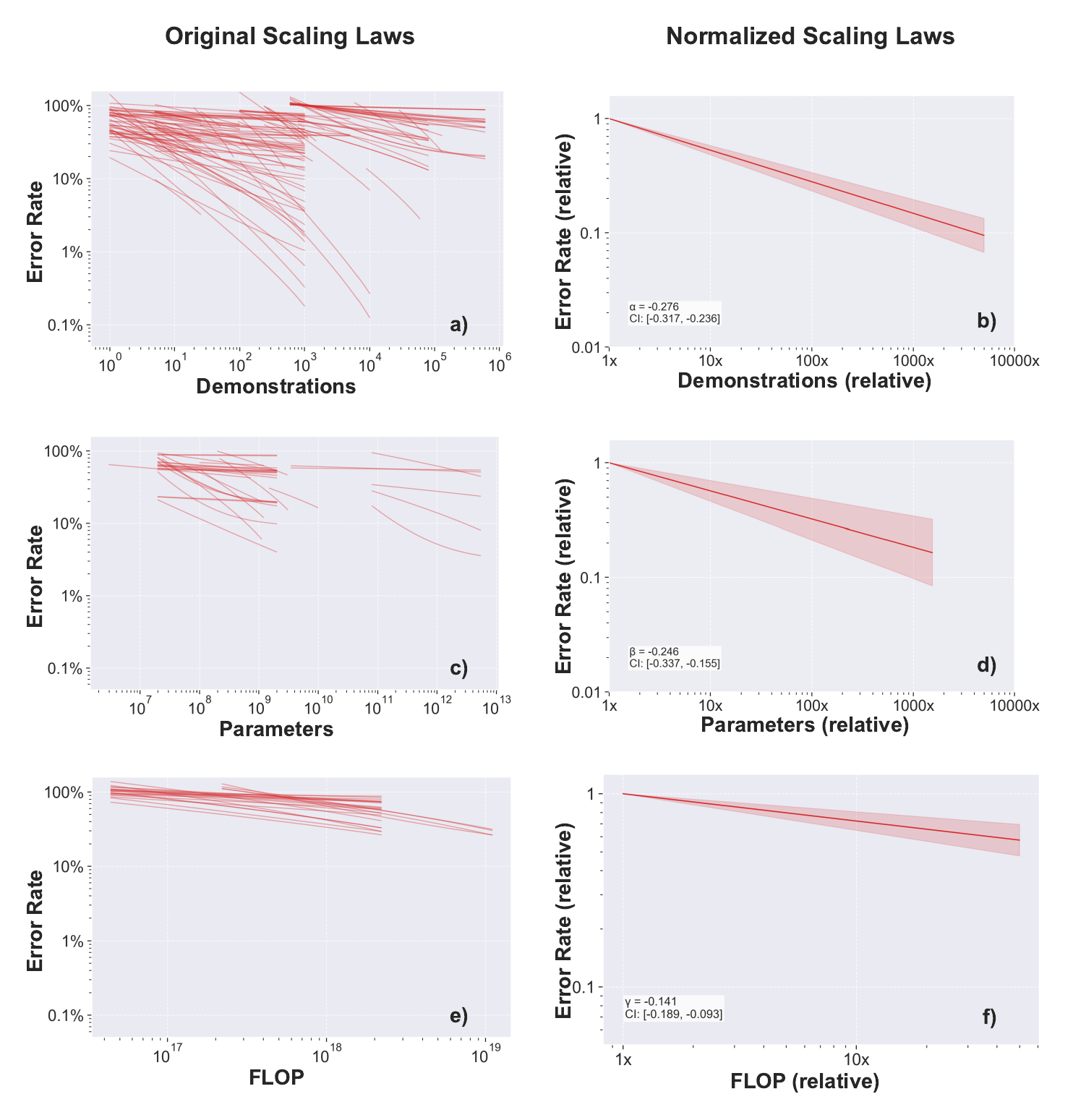}
    \caption{Scaling laws in robotics: (a, c, e) show scaling across data, model size, and compute, with fitted power laws. (b, d, f) illustrates the relative scaling behaviors as \(D\), \(N\), and \(C\) increase, modeled by \( \text{Error Rate}(X) = X^{(\bar{\sigma_X} \pm K_X)} \), where \( X \) denotes the input variable (\(D\), \(N\), or \(C\)), \( \bar{\sigma_X} \) is the mean scaling exponent (\( \alpha \),~\( \beta \),~or~\( \gamma \)), and \( K_X \) represents the Confidence Interval of \( \sigma_X \).}

    \label{fig:Scaling_Laws_All}
\end{figure}

Our findings indicate that the performance of robotics models generally improves with increased data, following a power-law relationship. This scaling behavior is illustrated in Figure~\ref{fig:Scaling_Laws_All}, while Table~\ref{tab: R_squared} compares the $R^2$ values for our fitted power-laws and linear models ($ErrorRate(X) = AX + B$). \( X \) denotes the input variable (\(D\), \(N\), or \(C\)) and \(A\) and \(B\) are constants. The results demonstrate that power-law models provide a far superior fit to the data. In Figure~\ref{fig:Scaling_Laws_All} the solid line represents the mean scaling law power exponent (\(\alpha\), \(\beta\), and \(\gamma\)), accompanied by a 95\% confidence interval. The \enquote{1x} marker denotes the smallest number of demonstrations used in the authors' scaling studies, while \enquote{10x}, \enquote{100x}, and beyond represent extrapolations based on the minimum delta between the maximum and minimum number of demonstrations across all studies. This visualization highlights the variation in power law exponent coefficients, reflecting differences in task difficulty and architecture differences. Notably, this pattern parallels findings in image generation, where image size influences distinct scaling laws~\citep{henighan2020scaling}.

The data scaling studies analyzed span datasets ranging from 1 to 1 million demonstrations, with a median ratio of 10x between the maximum and minimum number of demonstrations across all scaling law studies (see Table~\ref{tab:scaling_studies_stat}).

\begin{table}[t]
\caption{Explanatory power of power laws compared to linear models}
\label{tab: R_squared}
\centering
\begin{tabular}{lp{3cm}p{3cm}}
\toprule
\textbf{} & \textbf{$R^2$ power law} & \textbf{$R^2$ linear model} \\
\textbf{} & \textbf{Median (SD)} & \textbf{Median (SD)} \\
\midrule
\textbf{Data Size} & 0.93 (0.20) & 0.60 (0.30) \\
\textbf{Model Size} & 0.86 (0.28) & 0.64 (0.32) \\
\textbf{Compute} & 0.86 (0.18) & 0.72 (0.23) \\
\bottomrule
\end{tabular}
\end{table}

\begin{table}[ht]
\caption{Statistics on Scaling Studies}
\label{tab:scaling_studies_stat}
\centering
\begin{tabular}{lp{3.5cm}p{2.5cm}}
\toprule
\textbf{Metric} & \textbf{Overall range (in Orders of Magnitude)} & \textbf{Median Ratio Max/ Min Scaling Studies} \\
\midrule
\textbf{Data Size} & 1 to 1.0M (6.0) & 10.0x \\
\textbf{Model Size} & 3M to 1.8T (6.8) & 10.3x \\
\textbf{Compute} & \(4.4 \times 10^{16}\) to \(1.1 \times 10^{19}\) (2.4) & 50.0x \\
\bottomrule
\end{tabular}
\end{table}

While we confirm the power law behavior, this result is consistent with prior studies across various domains that have established and validated the ubiquity of power law scaling. However, it is important to emphasize that power laws inherently exhibit diminishing returns as resources increase. This implies that achieving high accuracy and reliability requires disproportionately large amounts of resources, posing challenges due to data limitations. Furthermore, real-world industrial and home applications often demand accuracy and reliability levels of 99.X\% or higher. Given the long-tail nature of power law scaling, reaching such levels is particularly challenging~\citep{ScalingSolveRobotics}. Across our dataset of 424 studies (comprising 214 studies with at least three data points per scaling study, plus additional studies with two data points), the mean top performance achieved was 67\% (median 70\%; SD = 24\%), highlighting the need for more extensive research to achieve the desired levels of accuracy and reliability.

\begin{figure}[!ht]
        \includegraphics[width=0.5\textwidth]{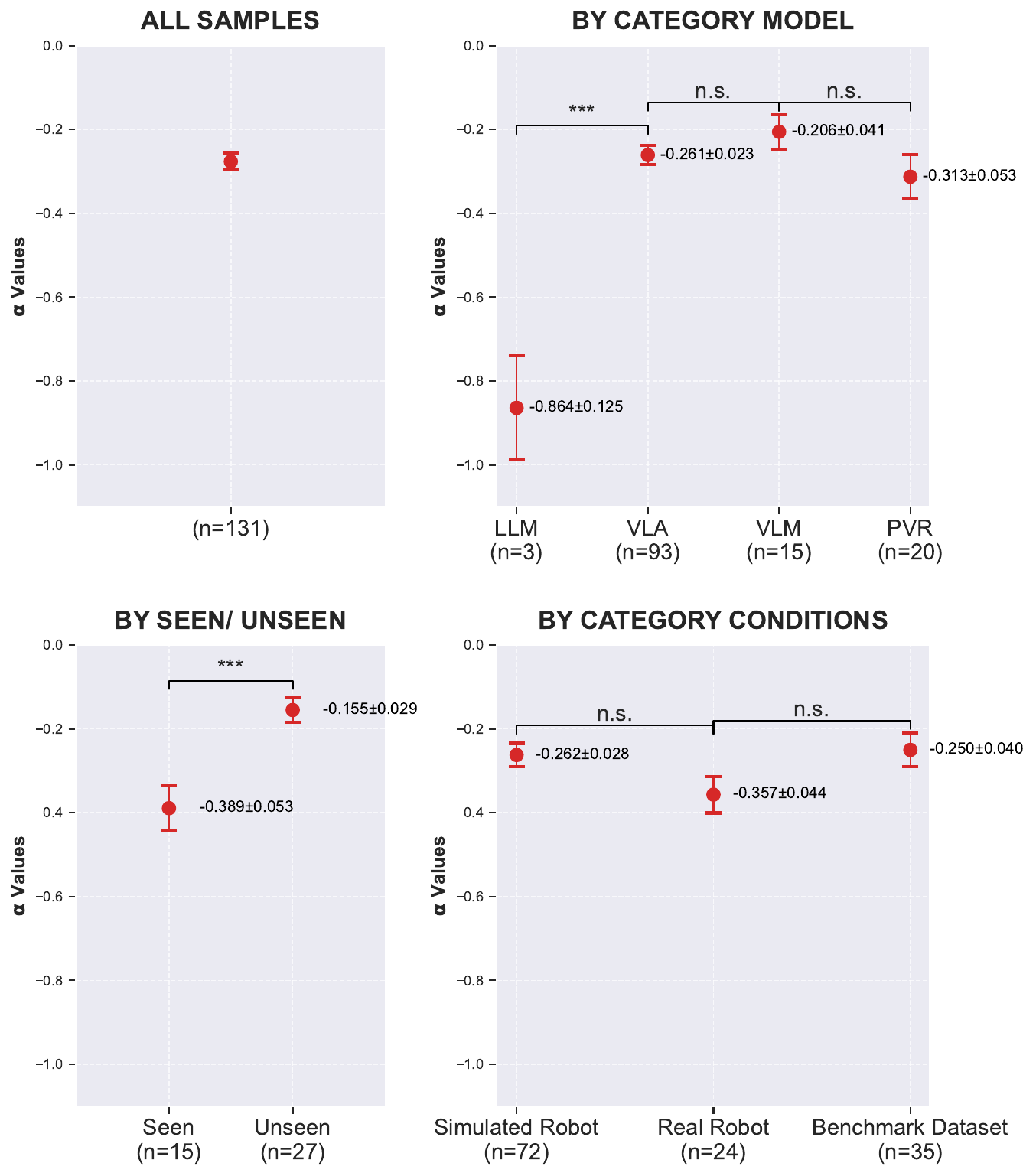}
        \caption{Data size scaling laws with error bars (mean, standard error). Statistical significance is indicated as follows: n.s.\ (not significant) $p > 0.05$; * $p \leq 0.05$; ** $p \leq 0.01$; *** $p \leq 0.001$.}
        \label{fig:Scaling_Laws_Data}
\end{figure}

Examining scaling efficiency (as shown in Figure~\ref{fig:Scaling_Laws_Data}), we find an average power law gradient (exponent) of -0.276 (95\% CI = -0.317, -0.236) across 131 data-scaling studies, where CI refers to the confidence interval. Most models analyzed are VLA models (n=93). Interestingly, while the mean \(\alpha\) value shows little variation between VLA models, VLMs, and PVRs, LLMs appear to scale significantly more efficiently. However, this observation should be treated with caution due to the small sample size.

\subsection{Scaling Laws: Model Size}
In comparison to data scaling, we observe fewer studies focused on model size (n=34). Nevertheless, similar to data, model performance improves with increasing model size and adheres to a power law relationship (see Figure~\ref{fig:Scaling_Laws_All} and Table~\ref{tab: R_squared}). The model sizes in these studies range from millions to trillions of parameters (refer to Table~\ref{tab:scaling_studies_stat}).

\begin{figure}[!ht]
        \includegraphics[width=0.5\textwidth]{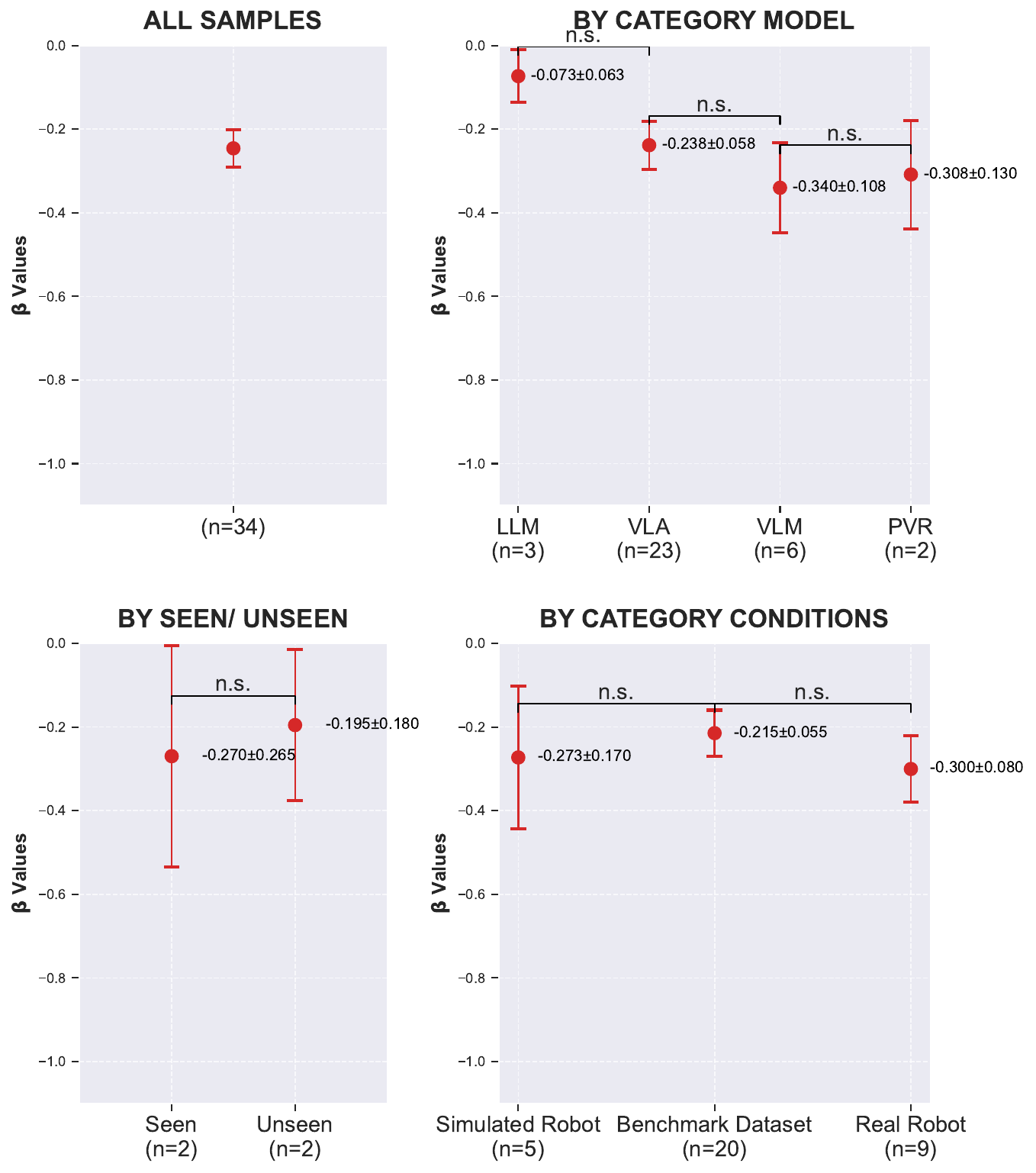}
        \caption{Model size scaling laws with error bars (mean, standard error). Statistical significance is indicated as follows: n.s.\ (not significant) $p > 0.05$; * $p \leq 0.05$; ** $p \leq 0.01$; *** $p \leq 0.001$.}
        \label{fig:Scaling_Laws_Model}
\end{figure}

The average power law gradient for the scaling exponent is -0.246 (95\% CI = -0.337, -0.155) across all 34 model size scaling studies. Beyond this, the insights largely align with those observed in data scaling. While some deviations exist (e.g., for LLMs), these are likely attributable to the low statistical power of the findings, as only three scaling studies per model type were available (see Figure~\ref{fig:Scaling_Laws_Model}).

\subsection{Scaling Laws: Compute}
Compared to data and model size, compute scaling has received considerably less attention in robotics research within our dataset. Specifically, we identified only one study that investigated the impact of compute scaling on performance, spanning 6 tasks and 4 architectures, resulting in 24 scaling studies in total~\citep{liu2023tail}. While the limited number of studies on compute scaling is unfortunate, it offers valuable insights into how task and architecture influence scaling efficiency.

\begin{figure}[!ht]
        \includegraphics[width=0.5\textwidth]{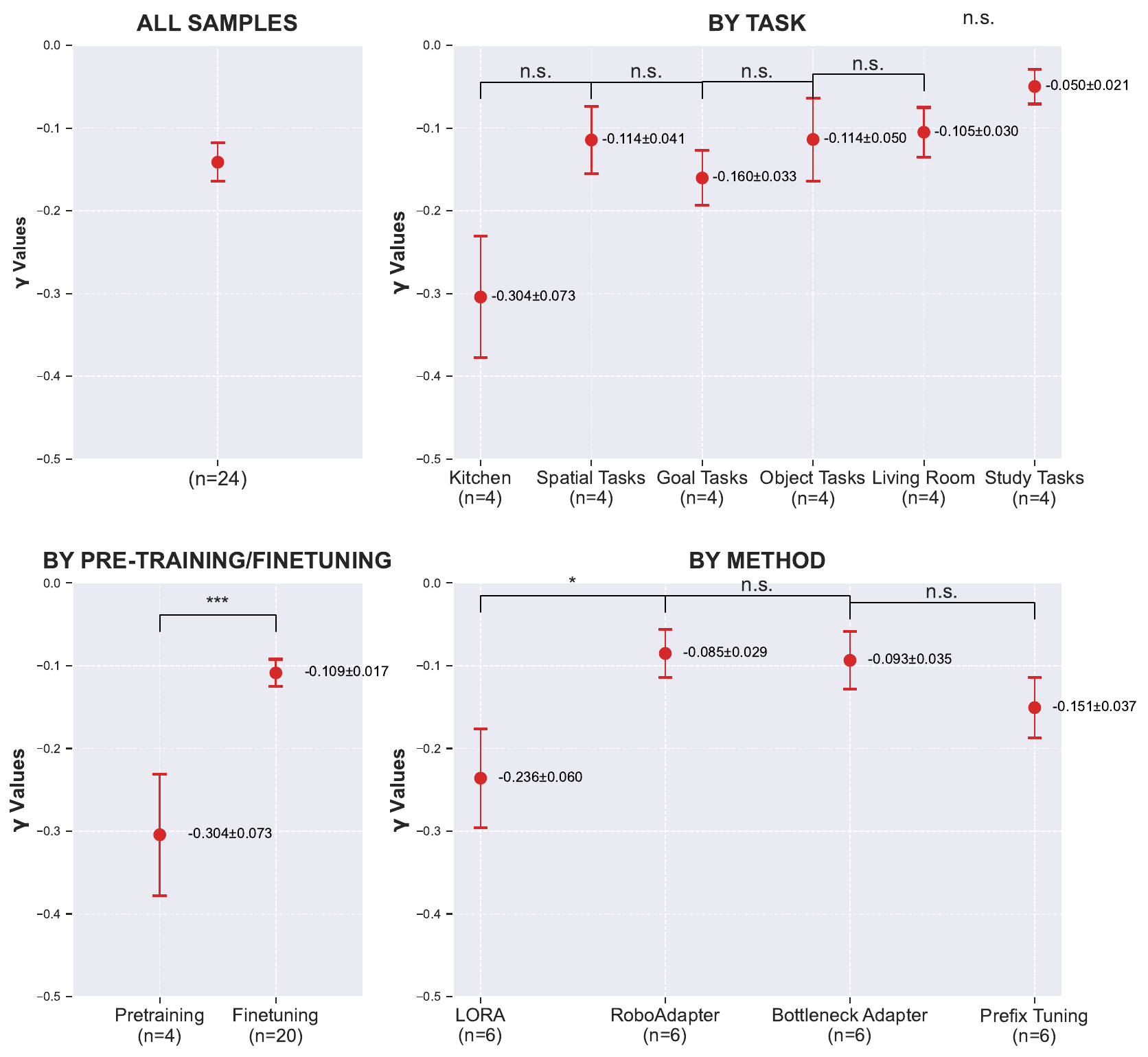}
        \caption{Compute scaling laws with error bars (mean, standard error). Statistical significance is indicated as follows: n.s.\ (not significant) $p > 0.05$; * $p \leq 0.05$; ** $p \leq 0.01$; *** $p \leq 0.001$.}
        \label{fig:Scaling_Laws_Compute}
\end{figure}

Our analysis reveals that the mean power law fit gradient for the scaling exponent ranges from -0.050 (95\% CI = -0.117, 0.017) to -0.304 (95\% CI = -0.538, -0.070), with an average of -0.141 (95\% CI = -0.189, -0.093) across all compute scaling studies. Interestingly, we observe differences in scaling efficiency between tasks and across different methods. This suggests that both architectural design and task characteristics significantly influence scaling efficiency, aligning with previous research by~\citet{rosenfeld2021predictability} and contrasting with findings from~\citet{hestness2017deep}. For example, variations in power coefficients have been reported for image generation tasks with differing pixel sizes (see Table~\ref{tab:scaling_laws_henighan}), underscoring the complexity and diversity of scaling behaviors. Additionally, our analysis finds that pretraining is generally more scaling-efficient than fine-tuning (Figure~\ref{fig:Scaling_Laws_Compute}).

The origins of scaling laws research in Natural Language Processing and other domains have traditionally placed a strong emphasis on compute. In robotics, however, data scaling appears to take higher priority, likely due to the absence of internet-scale datasets for embodied AI. This contrast highlights a fundamental difference between the two fields.

\subsection{Summary of Scaling Law Findings: Data, Model Size, and Compute}
The power coefficients \(\alpha\), \(\beta\) and \(\gamma\) are the most significant factors in understanding scaling behaviors, as outlined in Tables~\ref{tab:research_results_beta_all} and ~\ref{tab:research_results_beta_above_07}. While Table~\ref{tab:research_results_beta_all} presents all collected data, Table ~\ref{tab:research_results_beta_above_07} filters out outliers with low \(R^2\) values, focusing on data points where increased resources directly correlate with improved performance. This filtering approach reduces deviations between the median and mean values, refining our understanding of scaling efficiency.

For data size, the power coefficients range from -0.217 (median) to -0.276 (mean); for model size, they span from -0.172 to -0.246, and for compute, from -0.105 to -0.141. This means, that that doubling performance in robotics requires scaling data by $24.39\times$, parameters by $56.26\times$, and compute by $736.13\times$ (median power exponent used for calculation).

Interestingly, scaling coefficients for seen data (mean -0.389; 95\% CI: -0.502, -0.276) substantially outperform those for unseen data (mean -0.155; 95\% CI: -0.216, -0.094), reinforcing the importance of diverse and comprehensive datasets in improving general-purpose robotics performance.

Our analysis indicates that 87\% of scaling studies show \(\alpha\), \(\beta\) and \(\gamma\) values between 0 and -1, suggesting diminishing returns as resources increase. Considering the constraints of compute, data availability in robotics (with the largest dataset comprising 2.5 million episodes~\citep{padalkar2023open}), and the challenges of inference on edge devices, achieving efficient scaling in robotics remains a significant challenge.

These findings broadly align with prior studies on scaling laws in robotics and embodied AI, though some differences in the observed \(\alpha\), \(\beta\) and \(\gamma\) values are notable. Specifically, we observe power exponent values for data size ranging from -0.01 to -1, with a mean of -0.276 (131 scaling studies). In comparison, \citet{lin2024data} reports a more constrained range from -0.446 to -0.844 across tasks (6 scaling studies). Our compute scaling laws span from -0.01 to -0.52 (24 scaling studies), which is somewhat broader than the range of -0.03 to -0.31 (mean of -0.141) reported by~\citet{pearce2024scaling} (4 scaling studies). Despite these variations, the overall trend of diminishing returns as resources increase remains consistent with previous studies, reinforcing the challenges of scaling in robotics and embodied AI.

\begin{table}[!ht]
\caption{\(\alpha\), \(\beta\) and \(\gamma\) of power law approximation of Robot Foundation Models (RFMs) and LLMs in robotics}
\label{tab:research_results_beta_all}
\centering
\footnotesize
\setlength{\tabcolsep}{5pt}
\begin{tabular}{lccc}
\toprule
 &  & \textbf{\(\alpha\), \(\beta\) and \(\gamma\)} & \textbf{\(\alpha\), \(\beta\) and \(\gamma\)} \\
 & \textbf{n} & \textbf{Mean (Median)} & \textbf{CI (95\%)} \\
Data all samples & 131 & -0.276 (-0.217) & (-0.317, -0.236) \\
Data LLM & 3 & -0.864 (-0.978) & (-1.401, -0.328) \\
Data VLA & 93 & -0.261 (-0.225) & (-0.306, -0.216) \\
Data VLM & 15 & -0.206 (-0.156) & (-0.294, -0.117) \\
Data PVR & 20 & -0.313 (-0.251) & (-0.423, -0.202) \\
Data Unseen & 27 & -0.155 (-0.113) & (-0.216, -0.094) \\
Data Seen & 15 & -0.389 (-0.377) & (-0.502, -0.276) \\
Data Simulated Robot & 72 & -0.262 (-0.221) & (-0.317, -0.207) \\
Data Real Robot & 24 & -0.357 (-0.354) & (-0.447, -0.266) \\
Data Benchmark & 35 & -0.250 (-0.156) & (-0.332, -0.169) \\
\midrule
Model Size all samples & 34 & -0.246 (-0.172) & (-0.337, -0.155) \\
Model Size LLM & 3 & -0.073 (-0.015) & (-0.345, 0.199) \\
Model Size VLA & 23 & -0.238 (-0.135) & (-0.358, -0.119) \\
Model Size VLM & 6 & -0.340 (-0.297) & (-0.617, -0.063) \\
Model Size PVR & 2 & -0.308 (-0.308) & (-1.956, 1.339) \\
Model Size Unseen & 2 & -0.195 (-0.195) & (-2.487, 2.096) \\
Model Size Seen & 2 & -0.270 (-0.270) & (-3.633, 3.094) \\
Model Size Simulated Robot & 5 & -0.273 (-0.019) & (-0.745, 0.200) \\
Model Size Real Robot & 9 & -0.300 (-0.219) & (-0.484, -0.116) \\
Model Size Benchmark & 20 & -0.215 (-0.150) & (-0.330, -0.099) \\
\midrule
Compute all samples & 24 & -0.141 (-0.105) & (-0.189, -0.093) \\
Compute Kitchen & 4 & -0.304 (-0.242) & (-0.538, -0.070) \\
Compute Spatial Tasks & 4 & -0.114 (-0.081) & (-0.244, 0.015) \\
Compute Goals Tasks & 4 & -0.160 (-0.150) & (-0.266, -0.055) \\
Compute Object Tasks & 4 & -0.114 (-0.108) & (-0.273, 0.046) \\
Compute Living Room & 4 & -0.105 (-0.086) & (-0.200, -0.010) \\
Compute Study Tasks & 4 & -0.050 (-0.039) & (-0.117, 0.017) \\
Compute Pretraining & 4 & -0.304 (-0.242) & (-0.538, -0.070) \\
Compute Finetuning & 20 & -0.109 (-0.092) & (-0.143, -0.074) \\
Compute LORA & 6 & -0.236 (-0.188) & (-0.390, -0.082) \\
Compute RoboAdapter & 6 & -0.085 (-0.059) & (-0.160, -0.010) \\
Compute Bottleneck Adapter & 6 & -0.093 (-0.083) & (-0.183, -0.004) \\
Compute Prefix Tuning & 6 & -0.151 (-0.153) & (-0.245, -0.056) \\
\bottomrule
\end{tabular}
\end{table}

\begin{table}[!ht]
\caption{\(\alpha\), \(\beta\) and \(\gamma\) of power law approximation of Robot Foundation Models (RFMs) and LLMs in robotics for $R^2 > 0.7$}
\label{tab:research_results_beta_above_07}
\centering
\footnotesize
\setlength{\tabcolsep}{5pt}
\begin{tabular}{lccc}
\toprule
 &  & \textbf{\(\alpha\), \(\beta\) and \(\gamma\)} & \textbf{\(\alpha\), \(\beta\) and \(\gamma\)} \\
 & \textbf{n} & \textbf{Mean (Median)} & \textbf{CI (95\%)} \\
Data all samples & 115 & -0.299 (-0.256) & (-0.341, -0.256) \\
Data LLM & 3 & -0.864 (-0.978) & (-1.401, -0.328) \\
Data VLA & 77 & -0.291 (-0.278) & (-0.340, -0.243) \\
Data VLM & 15 & -0.206 (-0.156) & (-0.294, -0.117) \\
Data PVR & 20 & -0.313 (-0.251) & (-0.423, -0.202) \\
Data Unseen & 18 & -0.215 (-0.231) & (-0.291, -0.139) \\
Data Seen & 15 & -0.389 (-0.377) & (-0.502, -0.276) \\
Data Simulated Robot & 61 & -0.304 (-0.277) & (-0.363, -0.244) \\
Data Real Robot & 21 & -0.362 (-0.361) & (-0.456, -0.267) \\
Data Benchmark & 33 & -0.250 (-0.156) & (-0.334, -0.166) \\
\midrule
Model Size all samples & 23 & -0.331 (-0.244) & (-0.449, -0.212) \\
Model Size LLM & 3 & -0.073 (-0.015) & (-0.345, 0.199) \\
Model Size VLA & 13 & -0.368 (-0.318) & (-0.552, -0.184) \\
Model Size VLM & 5 & -0.397 (-0.376) & (-0.708, -0.087) \\
Model Size PVR & 2 & -0.308 (-0.308) & (-1.956, 1.339) \\
Model Size Unseen & 2 & -0.195 (-0.195) & (-2.487, 2.096) \\
Model Size Seen & 2 & -0.270 (-0.270) & (-3.633, 3.094) \\
Model Size Simulated Robot & 4 & -0.336 (-0.243) & (-0.985, 0.313) \\
Model Size Real Robot & 8 & -0.331 (-0.297) & (-0.528, -0.134) \\
Model Size Benchmark & 11 & -0.328 (-0.244) & (-0.515, -0.142) \\
\midrule
Compute all samples & 20 & -0.161 (-0.150) & (-0.214, -0.109) \\
Compute Kitchen & 4 & -0.304 (-0.242) & (-0.538, -0.070) \\
Compute Spatial Tasks & 4 & -0.114 (-0.081) & (-0.244, 0.015) \\
Compute Goals Tasks & 3 & -0.183 (-0.173) & (-0.332, -0.033) \\
Compute Object Tasks & 2 & -0.200 (-0.200) & (-0.397, -0.003) \\
Compute Living Room & 4 & -0.105 (-0.086) & (-0.200, -0.010) \\
Compute Study Tasks & 3 & -0.062 (-0.048) & (-0.165, 0.040) \\
Compute Pretraining & 4 & -0.304 (-0.242) & (-0.538, -0.070) \\
Compute Finetuning & 16 & -0.126 (-0.105) & (-0.164, -0.088) \\
Compute LORA & 6 & -0.236 (-0.188) & (-0.390, -0.082) \\
Compute RoboAdapter & 5 & -0.097 (-0.060) & (-0.188, -0.007) \\
Compute Bottleneck Adapter & 3 & -0.141 (-0.101) & (-0.382, 0.099) \\
Compute Prefix Tuning & 6 & -0.151 (-0.153) & (-0.245, -0.056) \\
\bottomrule
\end{tabular}
\end{table}

\subsection{Comparison of Robotics Scaling Laws to Other Modalities}
Robotics scaling laws reveal \(\beta\) values similar to those observed in image generation and text-to-image models, across both model size, data, and compute metrics, as detailed in Tables~\ref{tab:scaling_laws_kaplan} and~\ref{tab:scaling_laws_henighan}. Intriguingly, language training (classical LLM) remains more resource-intensive, even though robotics has traditionally been viewed as one of the most demanding domains in AI. This discrepancy may be attributed to the extensive reliance on image and video data in robotics. Future research should explore how varying data types (e.g., language, image, video, and action) and their respective proportions impact scaling efficiencies.

\subsection{Emergent Capabilities of Robot Foundation Models and LLMs used in robotics}
\begin{figure}
    \centering
    \begin{subfigure}[b]{0.49\textwidth}
        \includegraphics[width=\textwidth]{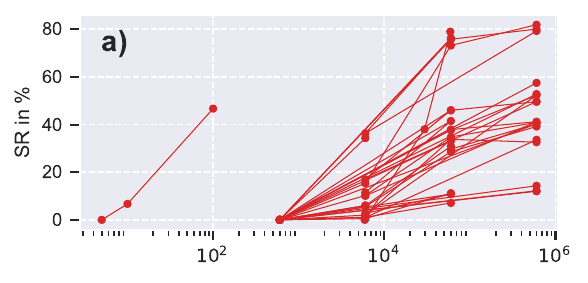}
        \label{fig:Emergent_D}
    \end{subfigure}
    \begin{subfigure}[b]{0.49\textwidth}
        \includegraphics[width=\textwidth]{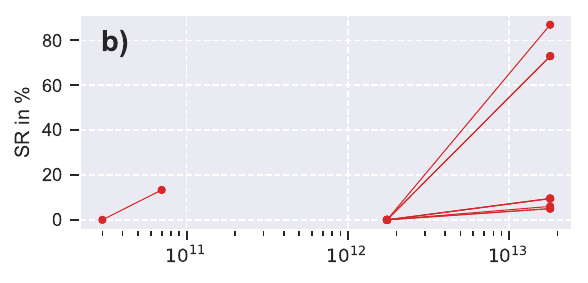}
        \label{fig:Emergent_N}
    \end{subfigure}
    \caption{Emergent Capabilities in Robotics: We find a wide range of emergent capabilities in robotics for both data (a) and model size (b).}
    \label{fig:Emerging}
\end{figure}

In this study, we observed cases of emergent behavior, visualized in Figure~\ref{fig:Emerging}. Previous work (e.g., \citep{chen2023autotamp, yang2023octopus, rana2023sayplan, stone2023open, bu2024synergisticgeneralizedefficientdualsystem, jiang2023vima}) has documented instances where models exhibited a 0\% success rate in their smallest-scale ablation studies but developed capabilities with increased scale.

For example, we noted emergent capabilities in task and motion planning for LLMs in robotics. However, these findings were inconsistently replicated across studies, and no significant patterns emerged due to the limited number of cases available for analysis. For model size, 10 cases of emergent behavior were observed across 125 scaling studies. Similarly, for data size, 18 out of 275 scaling studies exhibited emergent capabilities.

One reason for this inconsistency might be that researchers often prioritize reporting tasks where models succeed, leaving failures—such as cases with 0\% success—underreported. Yet, tasks with zero initial success rate are essential for identifying emergent capabilities. Future research should explicitly report tasks where models fail, aiding the study of emergent behavior and advancing our understanding of generalization.

\section{Discussion}
\paragraph{Summary} We conduct a meta-analysis on neural scaling laws in robotics, focusing on RFMs and LLMs in robotics, analyzed 327 papers to examine performance scaling with data, model size, and compute. Results confirm that power laws best describe these relationships, indicating diminishing returns with increased scale. Observed scaling laws for RFMs align with those in vision domains, with variations by task complexity and architectural designs. Emergent capabilities in RFMs and LLMs suggest potential for task generalization, offering insights for optimizing efficient robotic systems.

\paragraph{Limitations} 
Most scaling studies in robotics present only a limited number of data points, which constrains the precise estimation of power-law coefficients and contributes to variability in reported results. Moreover, model performance is often evaluated on tasks of varying complexity, leading to significant deviations even for the same model across different tasks. This underscores the pressing need for widely accepted, general-purpose, and open-ended benchmarks in robotics, similar to ImageNet in computer vision.
In addition to benchmarks, the field should establish standardized success rate metrics, as current practices vary significantly, further complicating comparisons. Unlike other machine learning domains, scaling studies in robotics often use validation sets that overlap with the training set, increasing the risk of overfitting to specific conditions.
Another key distinction is that scaling laws for foundation models in robotics do not follow predictable patterns like Chinchilla scaling laws, which assume proportional scaling of data, compute, and model size. This divergence makes direct comparisons between studies more challenging. Finally, there is a lack of differentiation between various architectural designs, which could also impact scalability and performance.

\paragraph{Ethical Considerations} While meta-analysis methods are typically not directly associated with ethical considerations, we recognize their significance and highlight the implications of our work. As we scale up robot systems, safety and control are paramount concerns due to the potential for physical harm. Additionally, as these models become more powerful, the complexity and risks increase, making robust safety protocols essential. However, our study not only acknowledges these risks but also offers positive contributions by enabling predictions about future models and their capabilities, assisting in establishing effective safety standards, and helping society prepare for and adapt to these technological advancements~\citep{WhiteHouse}. Moreover, the environmental impact of scaling AI cannot be ignored due to the immense computational resources required, necessitating eco-friendly approaches~\citep{thompson2021deep}. More specifically, the increasing cost of model training necessitate a shift towards algorithmic efficiency over relying on exponential compute scaling~\citep{thompson2017economic, leiserson2020there, thompson2021decline, thompson2022importance}. Lastly, the societal impact of widespread deployment of capable robotic agents must be carefully considered. While such systems hold the potential for substantial benefits, their disruptive effects on job markets, social structures, and existing power dynamics could exacerbate inequalities if not managed responsibly.

\section*{Acknowledgments}
This project has greatly benefited from discussions, ideas, inspiration and feedback provided by Jonathan Rosenfeld, Ana Trisovic, Emanuele Del Sozzo, Daniel Zhao, Gabriel Filipe, Alexander Fogelson, Hans Gundlach, Harry Lyu, Zachary Brown, Leonard Meinzinger and Simon Bohnen. We are also thankful to Prof. Joachim Henkel for co-supervising an earlier version of this work.

\bibliographystyle{plainnat}
\bibliography{references}

\begin{thebibliography}{92}
\providecommand{\natexlab}[1]{#1}
\providecommand{\url}[1]{\texttt{#1}}
\expandafter\ifx\csname urlstyle\endcsname\relax
  \providecommand{\doi}[1]{doi: #1}\else
  \providecommand{\doi}{doi: \begingroup \urlstyle{rm}\Url}\fi

\bibitem[AgiBot(2025)]{AgibotWorld}
AgiBot.
\newblock \href{https://agibot-world.com/}{AgiBot World Dataset}, 2025.

\bibitem[Ahmed et~al.(2023)Ahmed, Wahed, and Thompson]{ahmed2023growing}
Nur Ahmed, Muntasir Wahed, and Neil~C Thompson.
\newblock \href{https://www.science.org/doi/10.1126/science.ade2420}{The growing influence of industry in AI research}.
\newblock \emph{Science}, 379\penalty0 (6635):\penalty0 884--886, 2023.

\bibitem[Alabdulmohsin et~al.(2022)Alabdulmohsin, Neyshabur, and Zhai]{alabdulmohsin2022revisiting}
Ibrahim~M Alabdulmohsin, Behnam Neyshabur, and Xiaohua Zhai.
\newblock \href{https://arxiv.org/pdf/2209.06640}{Revisiting neural scaling laws in language and vision}.
\newblock \emph{Advances in Neural Information Processing Systems}, 35:\penalty0 22300--22312, 2022.

\bibitem[Amodei and Hernandez(2018)]{OpenAI_compute}
Dario Amodei and Danny Hernandez.
\newblock \href{https://openai.com/index/ai-and-compute/}{AI and compute}, 2018.

\bibitem[Black et~al.(2024)Black, Brown, Driess, Esmail, Equi, Finn, Fusai, Groom, Hausman, Ichter, et~al.]{black2024pi_0}
Kevin Black, Noah Brown, Danny Driess, Adnan Esmail, Michael Equi, Chelsea Finn, Niccolo Fusai, Lachy Groom, Karol Hausman, Brian Ichter, et~al.
\newblock \href{https://arxiv.org/pdf/2410.24164}{$\pi\_0$: A Vision-Language-Action Flow Model for General Robot Control}.
\newblock \emph{arXiv preprint arXiv:2410.24164}, 2024.

\bibitem[Bommasani et~al.(2021)Bommasani, Hudson, Adeli, Altman, Arora, von Arx, Bernstein, Bohg, Bosselut, Brunskill, et~al.]{bommasani2021opportunities}
Rishi Bommasani, Drew~A Hudson, Ehsan Adeli, Russ Altman, Simran Arora, Sydney von Arx, Michael~S Bernstein, Jeannette Bohg, Antoine Bosselut, Emma Brunskill, et~al.
\newblock \href{https://arxiv.org/pdf/2108.07258}{On the opportunities and risks of foundation models}.
\newblock \emph{arXiv preprint arXiv:2108.07258}, 2021.

\bibitem[Bousmalis et~al.(2023)Bousmalis, Vezzani, Rao, Devin, Lee, Villalonga, Davchev, Zhou, Gupta, Raju, et~al.]{bousmalis2023robocat}
Konstantinos Bousmalis, Giulia Vezzani, Dushyant Rao, Coline~Manon Devin, Alex~X Lee, Maria~Bauza Villalonga, Todor Davchev, Yuxiang Zhou, Agrim Gupta, Akhil Raju, et~al.
\newblock \href{https://arxiv.org/pdf/2306.11706}{RoboCat: A Self-Improving Generalist Agent for Robotic Manipulation}.
\newblock \emph{Transactions on Machine Learning Research}, 2023.

\bibitem[Brohan et~al.(2022)Brohan, Brown, Carbajal, Chebotar, Dabis, Finn, Gopalakrishnan, Hausman, Herzog, Hsu, et~al.]{brohan2022rt}
Anthony Brohan, Noah Brown, Justice Carbajal, Yevgen Chebotar, Joseph Dabis, Chelsea Finn, Keerthana Gopalakrishnan, Karol Hausman, Alex Herzog, Jasmine Hsu, et~al.
\newblock \href{https://arxiv.org/pdf/2212.06817}{Rt-1: Robotics transformer for real-world control at scale}.
\newblock \emph{arXiv preprint arXiv:2212.06817}, 2022.

\bibitem[Brohan et~al.(2023)Brohan, Brown, Carbajal, Chebotar, Chen, Choromanski, Ding, Driess, Dubey, Finn, et~al.]{brohan2023rt}
Anthony Brohan, Noah Brown, Justice Carbajal, Yevgen Chebotar, Xi~Chen, Krzysztof Choromanski, Tianli Ding, Danny Driess, Avinava Dubey, Chelsea Finn, et~al.
\newblock \href{https://arxiv.org/pdf/2307.15818}{Rt-2: Vision-language-action models transfer web knowledge to robotic control}.
\newblock \emph{arXiv preprint arXiv:2307.15818}, 2023.

\bibitem[Bu et~al.(2024)Bu, Li, Chen, Cai, Zeng, Cui, Yao, and Qiao]{bu2024synergisticgeneralizedefficientdualsystem}
Qingwen Bu, Hongyang Li, Li~Chen, Jisong Cai, Jia Zeng, Heming Cui, Maoqing Yao, and Yu~Qiao.
\newblock \href{https://arxiv.org/pdf/2410.08001}{Towards Synergistic, Generalized, and Efficient Dual-System for Robotic Manipulation}, 2024.

\bibitem[Cheang et~al.(2024)Cheang, Chen, Jing, Kong, Li, Li, Liu, Wu, Xu, Yang, Zhang, and Zhu]{cheang2024gr2generativevideolanguageactionmodel}
Chi-Lam Cheang, Guangzeng Chen, Ya~Jing, Tao Kong, Hang Li, Yifeng Li, Yuxiao Liu, Hongtao Wu, Jiafeng Xu, Yichu Yang, Hanbo Zhang, and Minzhao Zhu.
\newblock \href{https://arxiv.org/pdf/2410.06158}{GR-2: A Generative Video-Language-Action Model with Web-Scale Knowledge for Robot Manipulation}, 2024.

\bibitem[Chen et~al.(2023{\natexlab{a}})Chen, Xia, Ichter, Rao, Gopalakrishnan, Ryoo, Stone, and Kappler]{chen2023open}
Boyuan Chen, Fei Xia, Brian Ichter, Kanishka Rao, Keerthana Gopalakrishnan, Michael~S Ryoo, Austin Stone, and Daniel Kappler.
\newblock \href{https://arxiv.org/pdf/2209.09874}{Open-vocabulary queryable scene representations for real world planning}.
\newblock In \emph{2023 IEEE International Conference on Robotics and Automation (ICRA)}, pages 11509--11522. IEEE, 2023{\natexlab{a}}.

\bibitem[Chen et~al.(2023{\natexlab{b}})Chen, Arkin, Zhang, Roy, and Fan]{chen2023autotamp}
Yongchao Chen, Jacob Arkin, Yang Zhang, Nicholas Roy, and Chuchu Fan.
\newblock \href{https://arxiv.org/pdf/2306.06531}{Autotamp: Autoregressive task and motion planning with llms as translators and checkers}.
\newblock \emph{arXiv preprint arXiv:2306.06531}, 2023{\natexlab{b}}.

\bibitem[Coates et~al.(2011)Coates, Ng, and Lee]{coates2011analysis}
Adam Coates, Andrew Ng, and Honglak Lee.
\newblock \href{https://proceedings.mlr.press/v15/coates11a.html}{An analysis of single-layer networks in unsupervised feature learning}.
\newblock In \emph{Proceedings of the fourteenth international conference on artificial intelligence and statistics}, pages 215--223. JMLR Workshop and Conference Proceedings, 2011.

\bibitem[DAIR.AI(2024)]{MLPaperofTheWeek}
DAIR.AI.
\newblock \href{https://github.com/dair-ai/ML-Papers-of-the-Week}{ML Papers of The Week}, 2024.

\bibitem[Driess et~al.(2023)Driess, Xia, Sajjadi, Lynch, Chowdhery, Ichter, Wahid, Tompson, Vuong, Yu, et~al.]{driess2023palm}
Danny Driess, Fei Xia, Mehdi~SM Sajjadi, Corey Lynch, Aakanksha Chowdhery, Brian Ichter, Ayzaan Wahid, Jonathan Tompson, Quan Vuong, Tianhe Yu, et~al.
\newblock \href{https://arxiv.org/pdf/2303.03378}{Palm-e: An embodied multimodal language model}.
\newblock \emph{arXiv preprint arXiv:2303.03378}, 2023.

\bibitem[Duan et~al.(2024)Duan, Pumacay, Kumar, Wang, Tian, Yuan, Krishna, Fox, Mandlekar, and Guo]{duan2024aha}
Jiafei Duan, Wilbert Pumacay, Nishanth Kumar, Yi~Ru Wang, Shulin Tian, Wentao Yuan, Ranjay Krishna, Dieter Fox, Ajay Mandlekar, and Yijie Guo.
\newblock \href{https://arxiv.org/pdf/2410.00371}{AHA: A Vision-Language-Model for Detecting and Reasoning Over Failures in Robotic Manipulation}.
\newblock \emph{arXiv preprint arXiv:2410.00371}, 2024.

\bibitem[Etukuru et~al.(2024)Etukuru, Naka, Hu, Lee, Mehu, Edsinger, Paxton, Chintala, Pinto, and Shafiullah]{etukuru2024robot}
Haritheja Etukuru, Norihito Naka, Zijin Hu, Seungjae Lee, Julian Mehu, Aaron Edsinger, Chris Paxton, Soumith Chintala, Lerrel Pinto, and Nur Muhammad~Mahi Shafiullah.
\newblock \href{https://arxiv.org/pdf/2409.05865}{Robot utility models: General policies for zero-shot deployment in new environments}.
\newblock \emph{arXiv preprint arXiv:2409.05865}, 2024.

\bibitem[Firoozi et~al.(2023)Firoozi, Tucker, Tian, Majumdar, Sun, Liu, Zhu, Song, Kapoor, Hausman, et~al.]{survey_firoozi2023foundation}
Roya Firoozi, Johnathan Tucker, Stephen Tian, Anirudha Majumdar, Jiankai Sun, Weiyu Liu, Yuke Zhu, Shuran Song, Ashish Kapoor, Karol Hausman, et~al.
\newblock \href{https://arxiv.org/pdf/2312.07843}{Foundation models in robotics: Applications, challenges, and the future}.
\newblock \emph{arXiv preprint arXiv:2312.07843}, 2023.

\bibitem[Firoozi et~al.(2024)Firoozi, Tucker, Tian, Majumdar, Sun, Liu, Zhu, Song, Kapoor, Hausman, et~al.]{GitHub_Awesome-Robotics-Foundation-Models}
Roya Firoozi, Johnathan Tucker, Stephen Tian, Anirudha Majumdar, Jiankai Sun, Weiyu Liu, Yuke Zhu, Shuran Song, Ashish Kapoor, Karol Hausman, et~al.
\newblock \href{https://github.com/robotics-survey/Awesome-Robotics-Foundation-Models}{Awesome-Robotics-Foundation-Models}, 2024.

\bibitem[Fu et~al.(2024)Fu, Zhao, and Finn]{fu2024mobile}
Zipeng Fu, Tony~Z Zhao, and Chelsea Finn.
\newblock \href{https://arxiv.org/pdf/2401.02117}{Mobile aloha: Learning bimanual mobile manipulation with low-cost whole-body teleoperation}.
\newblock \emph{arXiv preprint arXiv:2401.02117}, 2024.

\bibitem[Ganguli et~al.(2022)Ganguli, Hernandez, Lovitt, Askell, Bai, Chen, Conerly, Dassarma, Drain, Elhage, et~al.]{ganguli2022predictability}
Deep Ganguli, Danny Hernandez, Liane Lovitt, Amanda Askell, Yuntao Bai, Anna Chen, Tom Conerly, Nova Dassarma, Dawn Drain, Nelson Elhage, et~al.
\newblock \href{https://arxiv.org/pdf/2202.07785}{Predictability and surprise in large generative models}.
\newblock In \emph{Proceedings of the 2022 ACM Conference on Fairness, Accountability, and Transparency}, pages 1747--1764, 2022.

\bibitem[Gao et~al.(2024)Gao, Xie, Xiao, Finn, and Sadigh]{gao2024efficientdatacollectionrobotic}
Jensen Gao, Annie Xie, Ted Xiao, Chelsea Finn, and Dorsa Sadigh.
\newblock \href{https://arxiv.org/pdf/2403.05110}{Efficient Data Collection for Robotic Manipulation via Compositional Generalization}, 2024.

\bibitem[Heim(2023)]{Heim_Compute}
Lennart Heim.
\newblock \href{https://blog.heim.xyz/flop-for-quantity-flop-s-for-performance/}{FLOP for Quantity, FLOP/s for Performance}, 2023.

\bibitem[Henighan et~al.(2020)Henighan, Kaplan, Katz, Chen, Hesse, Jackson, Jun, Brown, Dhariwal, Gray, et~al.]{henighan2020scaling}
Tom Henighan, Jared Kaplan, Mor Katz, Mark Chen, Christopher Hesse, Jacob Jackson, Heewoo Jun, Tom~B Brown, Prafulla Dhariwal, Scott Gray, et~al.
\newblock \href{https://arxiv.org/pdf/2010.14701}{Scaling laws for autoregressive generative modeling}.
\newblock \emph{arXiv preprint arXiv:2010.14701}, 2020.

\bibitem[Hestness et~al.(2017)Hestness, Narang, Ardalani, Diamos, Jun, Kianinejad, Patwary, Yang, and Zhou]{hestness2017deep}
Joel Hestness, Sharan Narang, Newsha Ardalani, Gregory Diamos, Heewoo Jun, Hassan Kianinejad, Md~Mostofa~Ali Patwary, Yang Yang, and Yanqi Zhou.
\newblock \href{https://arxiv.org/pdf/1712.00409}{Deep learning scaling is predictable, empirically}.
\newblock \emph{arXiv preprint arXiv:1712.00409}, 2017.

\bibitem[Hilton et~al.(2023)Hilton, Tang, and Schulman]{hilton2023scaling}
Jacob Hilton, Jie Tang, and John Schulman.
\newblock \href{https://arxiv.org/pdf/2301.13442}{Scaling laws for single-agent reinforcement learning}.
\newblock \emph{arXiv preprint arXiv:2301.13442}, 2023.

\bibitem[Hoffmann et~al.(2022)Hoffmann, Borgeaud, Mensch, Buchatskaya, Cai, Rutherford, Casas, Hendricks, Welbl, Clark, et~al.]{hoffmann2022training}
Jordan Hoffmann, Sebastian Borgeaud, Arthur Mensch, Elena Buchatskaya, Trevor Cai, Eliza Rutherford, Diego de~Las Casas, Lisa~Anne Hendricks, Johannes Welbl, Aidan Clark, et~al.
\newblock \href{https://arxiv.org/abs/2203.15556}{Training compute-optimal large language models}.
\newblock \emph{arXiv preprint arXiv:2203.15556}, 2022.

\bibitem[Hu(2024)]{GitHub_robotics-fm-survey}
Jeffrey Hu.
\newblock \href{https://github.com/JeffreyYH/robotics-fm-survey/tree/master}{robotics-fm-survey}, 2024.

\bibitem[Hu et~al.(2023)Hu, Xie, Jain, Francis, Patrikar, Keetha, Kim, Xie, Zhang, Zhao, et~al.]{hu2023toward}
Yafei Hu, Quanting Xie, Vidhi Jain, Jonathan Francis, Jay Patrikar, Nikhil Keetha, Seungchan Kim, Yaqi Xie, Tianyi Zhang, Zhibo Zhao, et~al.
\newblock \href{https://arxiv.org/pdf/2312.08782}{Toward general-purpose robots via foundation models: A survey and meta-analysis}.
\newblock \emph{arXiv preprint arXiv:2312.08782}, 2023.

\bibitem[Huang et~al.(2023)Huang, Jiang, Dong, Qiao, Gao, and Li]{huang2023instruct2act}
Siyuan Huang, Zhengkai Jiang, Hao Dong, Yu~Qiao, Peng Gao, and Hongsheng Li.
\newblock \href{https://arxiv.org/pdf/2305.11176}{Instruct2act: Mapping multi-modality instructions to robotic actions with large language model}.
\newblock \emph{arXiv preprint arXiv:2305.11176}, 2023.

\bibitem[Jang(2024)]{EricJang}
Eric Jang.
\newblock \href{https://www.youtube.com/watch?v=laeJn2-CBTk}{Data Engines for Humanoid Robots}, 2024.

\bibitem[Jiang et~al.(2023)Jiang, Gupta, Zhang, Wang, Dou, Chen, Fei-Fei, Anandkumar, Zhu, and Fan]{jiang2023vima}
Yunfan Jiang, Agrim Gupta, Zichen Zhang, Guanzhi Wang, Yongqiang Dou, Yanjun Chen, Li~Fei-Fei, Anima Anandkumar, Yuke Zhu, and Linxi Fan.
\newblock \href{https://arxiv.org/pdf/2210.03094}{Vima: Robot manipulation with multimodal prompts}, 2023.

\bibitem[Kaplan et~al.(2020)Kaplan, McCandlish, Henighan, Brown, Chess, Child, Gray, Radford, Wu, and Amodei]{kaplan2020scaling}
Jared Kaplan, Sam McCandlish, Tom Henighan, Tom~B Brown, Benjamin Chess, Rewon Child, Scott Gray, Alec Radford, Jeffrey Wu, and Dario Amodei.
\newblock \href{https://arxiv.org/pdf/2001.08361}{Scaling laws for neural language models}.
\newblock \emph{arXiv preprint arXiv:2001.08361}, 2020.

\bibitem[Kawaharazuka et~al.(2024)Kawaharazuka, Matsushima, Gambardella, Guo, Paxton, and Zeng]{Kawaharazuka_2024}
Kento Kawaharazuka, Tatsuya Matsushima, Andrew Gambardella, Jiaxian Guo, Chris Paxton, and Andy Zeng.
\newblock \href{http://dx.doi.org/10.1080/01691864.2024.2408593}real-world robot applications of foundation models: a review.
\newblock \emph{Advanced Robotics}, 38\penalty0 (18):\penalty0 1232–1254, September 2024.
\newblock ISSN 1568-5535.
\newblock \doi{10.1080/01691864.2024.2408593}.

\bibitem[Khazatsky et~al.(2024)Khazatsky, Pertsch, Nair, Balakrishna, Dasari, Karamcheti, Nasiriany, Srirama, Chen, Ellis, Fagan, Hejna, Itkina, Lepert, Ma, Miller, Wu, Belkhale, Dass, Ha, Jain, Lee, Lee, Memmel, Park, Radosavovic, Wang, Zhan, Black, Chi, Hatch, Lin, Lu, Mercat, Rehman, Sanketi, Sharma, Simpson, Vuong, Walke, Wulfe, Xiao, Yang, Yavary, Zhao, Agia, Baijal, Castro, Chen, Chen, Chung, Drake, Foster, Gao, Herrera, Heo, Hsu, Hu, Jackson, Le, Li, Lin, Lin, Ma, Maddukuri, Mirchandani, Morton, Nguyen, O'Neill, Scalise, Seale, Son, Tian, Tran, Wang, Wu, Xie, Yang, Yin, Zhang, Bastani, Berseth, Bohg, Goldberg, Gupta, Gupta, Jayaraman, Lim, Malik, Martín-Martín, Ramamoorthy, Sadigh, Song, Wu, Yip, Zhu, Kollar, Levine, and Finn]{khazatsky2024droid}
Alexander Khazatsky, Karl Pertsch, Suraj Nair, Ashwin Balakrishna, Sudeep Dasari, Siddharth Karamcheti, Soroush Nasiriany, Mohan~Kumar Srirama, Lawrence~Yunliang Chen, Kirsty Ellis, Peter~David Fagan, Joey Hejna, Masha Itkina, Marion Lepert, Yecheng~Jason Ma, Patrick~Tree Miller, Jimmy Wu, Suneel Belkhale, Shivin Dass, Huy Ha, Arhan Jain, Abraham Lee, Youngwoon Lee, Marius Memmel, Sungjae Park, Ilija Radosavovic, Kaiyuan Wang, Albert Zhan, Kevin Black, Cheng Chi, Kyle~Beltran Hatch, Shan Lin, Jingpei Lu, Jean Mercat, Abdul Rehman, Pannag~R Sanketi, Archit Sharma, Cody Simpson, Quan Vuong, Homer~Rich Walke, Blake Wulfe, Ted Xiao, Jonathan~Heewon Yang, Arefeh Yavary, Tony~Z. Zhao, Christopher Agia, Rohan Baijal, Mateo~Guaman Castro, Daphne Chen, Qiuyu Chen, Trinity Chung, Jaimyn Drake, Ethan~Paul Foster, Jensen Gao, David~Antonio Herrera, Minho Heo, Kyle Hsu, Jiaheng Hu, Donovon Jackson, Charlotte Le, Yunshuang Li, Kevin Lin, Roy Lin, Zehan Ma, Abhiram Maddukuri, Suvir Mirchandani, Daniel Morton, Tony Nguyen,
  Abigail O'Neill, Rosario Scalise, Derick Seale, Victor Son, Stephen Tian, Emi Tran, Andrew~E. Wang, Yilin Wu, Annie Xie, Jingyun Yang, Patrick Yin, Yunchu Zhang, Osbert Bastani, Glen Berseth, Jeannette Bohg, Ken Goldberg, Abhinav Gupta, Abhishek Gupta, Dinesh Jayaraman, Joseph~J Lim, Jitendra Malik, Roberto Martín-Martín, Subramanian Ramamoorthy, Dorsa Sadigh, Shuran Song, Jiajun Wu, Michael~C. Yip, Yuke Zhu, Thomas Kollar, Sergey Levine, and Chelsea Finn.
\newblock \href{https://arxiv.org/pdf/2403.12945}{DROID: A Large-Scale In-The-Wild Robot Manipulation Dataset}, 2024.

\bibitem[Kim et~al.(2024)Kim, Pertsch, Karamcheti, Xiao, Balakrishna, Nair, Rafailov, Foster, Lam, Sanketi, et~al.]{kim2024openvla}
Moo~Jin Kim, Karl Pertsch, Siddharth Karamcheti, Ted Xiao, Ashwin Balakrishna, Suraj Nair, Rafael Rafailov, Ethan Foster, Grace Lam, Pannag Sanketi, et~al.
\newblock \href{https://arxiv.org/pdf/2406.09246}{OpenVLA: An Open-Source Vision-Language-Action Model}.
\newblock \emph{arXiv preprint arXiv:2406.09246}, 2024.

\bibitem[Kuang et~al.(2024)Kuang, Ye, Geng, Mao, Deng, Guibas, Wang, and Wang]{kuang2024ramretrievalbasedaffordancetransfer}
Yuxuan Kuang, Junjie Ye, Haoran Geng, Jiageng Mao, Congyue Deng, Leonidas Guibas, He~Wang, and Yue Wang.
\newblock \href{https://arxiv.org/pdf/2407.04689}{RAM: Retrieval-Based Affordance Transfer for Generalizable Zero-Shot Robotic Manipulation}, 2024.

\bibitem[Kumar(2024)]{ScalingSolveRobotics}
Nishanth~J. Kumar.
\newblock \href{https://spectrum.ieee.org/solve-robotics}{Will Scaling Solve Robotics?}, 2024.

\bibitem[Leiserson et~al.(2020)Leiserson, Thompson, Emer, Kuszmaul, Lampson, Sanchez, and Schardl]{leiserson2020there}
Charles~E Leiserson, Neil~C Thompson, Joel~S Emer, Bradley~C Kuszmaul, Butler~W Lampson, Daniel Sanchez, and Tao~B Schardl.
\newblock \href{https://www.science.org/doi/10.1126/science.aam9744}{There’s plenty of room at the Top: What will drive computer performance after Moore’s law?}
\newblock \emph{Science}, 368\penalty0 (6495):\penalty0 eaam9744, 2020.

\bibitem[Li et~al.(2024{\natexlab{a}})Li, Jin, Yu, Shi, Hao, Hao, Liu, Sun, Zhang, Fang, et~al.]{li2024foundation}
Dingzhe Li, Yixiang Jin, Hongze Yu, Jun Shi, Xiaoshuai Hao, Peng Hao, Huaping Liu, Fuchun Sun, Jianwei Zhang, Bin Fang, et~al.
\newblock \href{https://arxiv.org/pdf/2404.18201}{What foundation models can bring for robot learning in manipulation: A survey}.
\newblock \emph{arXiv preprint arXiv:2404.18201}, 2024{\natexlab{a}}.

\bibitem[Li et~al.(2022)Li, Puig, Paxton, Du, Wang, Fan, Chen, Huang, Aky{\"u}rek, Anandkumar, et~al.]{li2022pre}
Shuang Li, Xavier Puig, Chris Paxton, Yilun Du, Clinton Wang, Linxi Fan, Tao Chen, De-An Huang, Ekin Aky{\"u}rek, Anima Anandkumar, et~al.
\newblock \href{https://arxiv.org/pdf/2202.01771}{Pre-trained language models for interactive decision-making}.
\newblock \emph{Advances in Neural Information Processing Systems}, 35:\penalty0 31199--31212, 2022.

\bibitem[Li et~al.(2024{\natexlab{b}})Li, Mata, Park, Kahatapitiya, Jang, Shang, Ranasinghe, Burgert, Cai, Lee, and Ryoo]{li2024llarasuperchargingrobotlearning}
Xiang Li, Cristina Mata, Jongwoo Park, Kumara Kahatapitiya, Yoo~Sung Jang, Jinghuan Shang, Kanchana Ranasinghe, Ryan Burgert, Mu~Cai, Yong~Jae Lee, and Michael~S. Ryoo.
\newblock \href{https://arxiv.org/pdf/2406.20095}{LLaRA: Supercharging Robot Learning Data for Vision-Language Policy}, 2024{\natexlab{b}}.

\bibitem[Lin et~al.(2024)Lin, Hu, Sheng, Wen, You, and Gao]{lin2024data}
Fanqi Lin, Yingdong Hu, Pingyue Sheng, Chuan Wen, Jiacheng You, and Yang Gao.
\newblock \href{https://arxiv.org/pdf/2410.18647}{Data scaling laws in imitation learning for robotic manipulation}.
\newblock \emph{arXiv preprint arXiv:2410.18647}, 2024.

\bibitem[Liu et~al.(2023)Liu, Zhang, Asadi, Liu, Zhao, Sabach, and Fakoor]{liu2023tail}
Zuxin Liu, Jesse Zhang, Kavosh Asadi, Yao Liu, Ding Zhao, Shoham Sabach, and Rasool Fakoor.
\newblock \href{https://arxiv.org/pdf/2310.05905}{TAIL: Task-specific Adapters for Imitation Learning with Large Pretrained Models}.
\newblock \emph{arXiv preprint arXiv:2310.05905}, 2023.

\bibitem[Lynch and Sermanet(2020)]{lynch2020language}
Corey Lynch and Pierre Sermanet.
\newblock \href{https://arxiv.org/pdf/2005.07648}{Language conditioned imitation learning over unstructured data}.
\newblock \emph{arXiv preprint arXiv:2005.07648}, 2020.

\bibitem[Majumdar et~al.(2024)Majumdar, Yadav, Arnaud, Ma, Chen, Silwal, Jain, Berges, Wu, Vakil, et~al.]{majumdar2024we}
Arjun Majumdar, Karmesh Yadav, Sergio Arnaud, Jason Ma, Claire Chen, Sneha Silwal, Aryan Jain, Vincent-Pierre Berges, Tingfan Wu, Jay Vakil, et~al.
\newblock \href{https://arxiv.org/pdf/2303.18240}{Where are we in the search for an artificial visual cortex for embodied intelligence?}
\newblock \emph{Advances in Neural Information Processing Systems}, 36, 2024.

\bibitem[Muennighoff et~al.(2024)Muennighoff, Rush, Barak, Le~Scao, Tazi, Piktus, Pyysalo, Wolf, and Raffel]{muennighoff2024scaling}
Niklas Muennighoff, Alexander Rush, Boaz Barak, Teven Le~Scao, Nouamane Tazi, Aleksandra Piktus, Sampo Pyysalo, Thomas Wolf, and Colin~A Raffel.
\newblock \href{https://arxiv.org/pdf/2305.16264}{Scaling data-constrained language models}.
\newblock \emph{Advances in Neural Information Processing Systems}, 36, 2024.

\bibitem[Nair et~al.(2022)Nair, Rajeswaran, Kumar, Finn, and Gupta]{nair2022r3m}
Suraj Nair, Aravind Rajeswaran, Vikash Kumar, Chelsea Finn, and Abhinav Gupta.
\newblock \href{https://arxiv.org/pdf/2203.12601}{R3m: A universal visual representation for robot manipulation}.
\newblock \emph{arXiv preprint arXiv:2203.12601}, 2022.

\bibitem[Nasiriany et~al.(2024)Nasiriany, Maddukuri, Zhang, Parikh, Lo, Joshi, Mandlekar, and Zhu]{robocasa2024}
Soroush Nasiriany, Abhiram Maddukuri, Lance Zhang, Adeet Parikh, Aaron Lo, Abhishek Joshi, Ajay Mandlekar, and Yuke Zhu.
\newblock \href{https://arxiv.org/pdf/2406.02523}{RoboCasa: Large-Scale Simulation of Everyday Tasks for Generalist Robots}.
\newblock In \emph{Robotics: Science and Systems (RSS)}, 2024.

\bibitem[Padalkar et~al.(2023)Padalkar, Pooley, Jain, Bewley, Herzog, Irpan, Khazatsky, Rai, Singh, Brohan, et~al.]{padalkar2023open}
Abhishek Padalkar, Acorn Pooley, Ajinkya Jain, Alex Bewley, Alex Herzog, Alex Irpan, Alexander Khazatsky, Anant Rai, Anikait Singh, Anthony Brohan, et~al.
\newblock \href{https://arxiv.org/pdf/2310.08864}{Open x-embodiment: Robotic learning datasets and rt-x models}.
\newblock \emph{arXiv preprint arXiv:2310.08864}, 2023.

\bibitem[Pearce et~al.(2024)Pearce, Rashid, Bignell, Georgescu, Devlin, and Hofmann]{pearce2024scaling}
Tim Pearce, Tabish Rashid, Dave Bignell, Raluca Georgescu, Sam Devlin, and Katja Hofmann.
\newblock \href{https://arxiv.org/pdf/2411.04434}{Scaling Laws for Pre-training Agents and World Models}.
\newblock \emph{arXiv preprint arXiv:2411.04434}, 2024.

\bibitem[Radosavovic et~al.(2023{\natexlab{a}})Radosavovic, Shi, Fu, Goldberg, Darrell, and Malik]{radosavovic2023robot}
Ilija Radosavovic, Baifeng Shi, Letian Fu, Ken Goldberg, Trevor Darrell, and Jitendra Malik.
\newblock \href{https://arxiv.org/pdf/2306.10007}{Robot learning with sensorimotor pre-training}.
\newblock In \emph{Conference on Robot Learning}, pages 683--693. PMLR, 2023{\natexlab{a}}.

\bibitem[Radosavovic et~al.(2023{\natexlab{b}})Radosavovic, Xiao, James, Abbeel, Malik, and Darrell]{radosavovic2023real}
Ilija Radosavovic, Tete Xiao, Stephen James, Pieter Abbeel, Jitendra Malik, and Trevor Darrell.
\newblock \href{https://arxiv.org/pdf/2210.03109}{Real-world robot learning with masked visual pre-training}.
\newblock In \emph{Conference on Robot Learning}, pages 416--426. PMLR, 2023{\natexlab{b}}.

\bibitem[Rana et~al.(2023)Rana, Haviland, Garg, Abou-Chakra, Reid, and Suenderhauf]{rana2023sayplan}
Krishan Rana, Jesse Haviland, Sourav Garg, Jad Abou-Chakra, Ian Reid, and Niko Suenderhauf.
\newblock \href{https://arxiv.org/pdf/2307.06135}{Sayplan: Grounding large language models using 3d scene graphs for scalable task planning}.
\newblock \emph{arXiv preprint arXiv:2307.06135}, 2023.

\bibitem[Rana et~al.(2024)Rana, Abou-Chakra, Garg, Lee, Reid, and Suenderhauf]{rana2024affordancecentricpolicylearningsample}
Krishan Rana, Jad Abou-Chakra, Sourav Garg, Robert Lee, Ian Reid, and Niko Suenderhauf.
\newblock \href{https://arxiv.org/pdf/2410.12124}{Affordance-Centric Policy Learning: Sample Efficient and Generalisable Robot Policy Learning using Affordance-Centric Task Frames}, 2024.

\bibitem[Reed et~al.(2022)Reed, Zolna, Parisotto, Colmenarejo, Novikov, Barth-Maron, Gimenez, Sulsky, Kay, Springenberg, et~al.]{reed2022generalist}
Scott Reed, Konrad Zolna, Emilio Parisotto, Sergio~Gomez Colmenarejo, Alexander Novikov, Gabriel Barth-Maron, Mai Gimenez, Yury Sulsky, Jackie Kay, Jost~Tobias Springenberg, et~al.
\newblock \href{https://arxiv.org/pdf/2205.06175}{A generalist agent}.
\newblock \emph{arXiv preprint arXiv:2205.06175}, 2022.

\bibitem[Rintamaki(2023)]{GitHub_Everything-LLMs-And-Robotics}
Jacob Rintamaki.
\newblock \href{https://github.com/jrin771/Everything-LLMs-And-Robotics}{Everything-LLMs-And-Robotics}, 2023.

\bibitem[Rosenfeld(2021)]{rosenfeld2021scaling}
Jonathan~S Rosenfeld.
\newblock \href{https://arxiv.org/pdf/2108.07686}{Scaling laws for deep learning}.
\newblock \emph{arXiv preprint arXiv:2108.07686}, 2021.

\bibitem[Rosenfeld et~al.(2019)Rosenfeld, Rosenfeld, Belinkov, and Shavit]{rosenfeld2019constructive}
Jonathan~S Rosenfeld, Amir Rosenfeld, Yonatan Belinkov, and Nir Shavit.
\newblock \href{https://arxiv.org/pdf/1909.12673}{A constructive prediction of the generalization error across scales}.
\newblock \emph{arXiv preprint arXiv:1909.12673}, 2019.

\bibitem[Rosenfeld et~al.(2021)Rosenfeld, Frankle, Carbin, and Shavit]{rosenfeld2021predictability}
Jonathan~S Rosenfeld, Jonathan Frankle, Michael Carbin, and Nir Shavit.
\newblock \href{https://arxiv.org/pdf/2006.10621}{On the predictability of pruning across scales}.
\newblock In \emph{International Conference on Machine Learning}, pages 9075--9083. PMLR, 2021.

\bibitem[Rosenfeld(2019)]{rosenfeld2019relation}
Jonathan~Shmuel Rosenfeld.
\newblock \emph{\href{https://dspace.mit.edu/handle/1721.1/122703}{On the relation between neural network size and performance}}.
\newblock PhD thesis, Massachusetts Institute of Technology, 2019.

\bibitem[Schaeffer et~al.(2024)Schaeffer, Miranda, and Koyejo]{schaeffer2024emergent}
Rylan Schaeffer, Brando Miranda, and Sanmi Koyejo.
\newblock \href{https://arxiv.org/pdf/2304.15004}{Are emergent abilities of large language models a mirage?}
\newblock \emph{Advances in Neural Information Processing Systems}, 36, 2024.

\bibitem[Sevilla et~al.(2022)Sevilla, Heim, Hobbhan, Besiroglu, Ho, and Villalobos]{Epoch_Compute}
Jaime Sevilla, Lennart Heim, Marius Hobbhan, Tamay Besiroglu, Anson Ho, and Pablo Villalobos.
\newblock \href{https://epochai.org/blog/estimating-training-compute}{Estimating Training Compute of Deep Learning Models}, 2022.

\bibitem[Sevilla et~al.(2023)Sevilla, Ho, and Besiroglu]{sevilla2023please}
Jaime Sevilla, Anson Ho, and Tamay Besiroglu.
\newblock \href{https://cacm.acm.org/opinion/please-report-your-compute/}{Please Report Your Compute}.
\newblock \emph{Communications of the ACM}, 66\penalty0 (5):\penalty0 30--32, 2023.

\bibitem[Shridhar et~al.(2022)Shridhar, Manuelli, and Fox]{shridhar2022cliport}
Mohit Shridhar, Lucas Manuelli, and Dieter Fox.
\newblock \href{https://arxiv.org/pdf/2109.12098}{Cliport: What and where pathways for robotic manipulation}.
\newblock In \emph{Conference on robot learning}, pages 894--906. PMLR, 2022.

\bibitem[Snell et~al.(2024)Snell, Lee, Xu, and Kumar]{snell2024scalingllmtesttimecompute}
Charlie Snell, Jaehoon Lee, Kelvin Xu, and Aviral Kumar.
\newblock \href{https://arxiv.org/pdf/2408.03314}{Scaling LLM Test-Time Compute Optimally can be More Effective than Scaling Model Parameters}, 2024.

\bibitem[Sorscher et~al.(2022)Sorscher, Geirhos, Shekhar, Ganguli, and Morcos]{sorscher2022beyond}
Ben Sorscher, Robert Geirhos, Shashank Shekhar, Surya Ganguli, and Ari Morcos.
\newblock \href{https://arxiv.org/pdf/2206.14486}{Beyond neural scaling laws: beating power law scaling via data pruning}.
\newblock \emph{Advances in Neural Information Processing Systems}, 35:\penalty0 19523--19536, 2022.

\bibitem[Staroverov et~al.(2023)Staroverov, Gorodetsky, Krishtopik, Izmesteva, Yudin, Kovalev, and Panov]{staroverov2023fine}
Aleksei Staroverov, Andrey~S Gorodetsky, Andrei~S Krishtopik, Uliana~A Izmesteva, Dmitry~A Yudin, Alexey~K Kovalev, and Aleksandr~I Panov.
\newblock \href{https://ieeexplore.ieee.org/document/10323309}{Fine-Tuning Multimodal Transformer Models for Generating Actions in Virtual and Real Environments}.
\newblock \emph{IEEE Access}, 11:\penalty0 130548--130559, 2023.

\bibitem[Stone et~al.(2023)Stone, Xiao, Lu, Gopalakrishnan, Lee, Vuong, Wohlhart, Kirmani, Zitkovich, Xia, et~al.]{stone2023open}
Austin Stone, Ted Xiao, Yao Lu, Keerthana Gopalakrishnan, Kuang-Huei Lee, Quan Vuong, Paul Wohlhart, Sean Kirmani, Brianna Zitkovich, Fei Xia, et~al.
\newblock \href{https://arxiv.org/pdf/2303.00905}{Open-world object manipulation using pre-trained vision-language models}.
\newblock \emph{arXiv preprint arXiv:2303.00905}, 2023.

\bibitem[Sutton(2019)]{sutton2019bitter}
Richard Sutton.
\newblock \href{https://www.cs.utexas.edu/~eunsol/courses/data/bitter_lesson.pdf}{The bitter lesson}.
\newblock \emph{Incomplete Ideas (blog)}, 13\penalty0 (1):\penalty0 38, 2019.

\bibitem[Tang et~al.(2024)Tang, Pan, Zhan, Zhou, Yao, Liu, Tomizuka, Ding, and Fu]{tang2024embodimentagnosticactionplanningobjectpart}
Weiliang Tang, Jia-Hui Pan, Wei Zhan, Jianshu Zhou, Huaxiu Yao, Yun-Hui Liu, Masayoshi Tomizuka, Mingyu Ding, and Chi-Wing Fu.
\newblock \href{https://www.arxiv.org/pdf/2409.10032}{Embodiment-Agnostic Action Planning via Object-Part Scene Flow}, 2024.

\bibitem[The~White~House(2023)]{WhiteHouse}
Biden The~White~House.
\newblock \href{https://www.whitehouse.gov/briefing-room/presidential-actions/2023/10/30/executive-order-on-the-safe-secure-and-trustworthy-development-and-use-of-artificial-intelligence/}{Executive Order on the Safe, Secure, and Trustworthy Development and Use of Artificial Intelligence}, 2023.

\bibitem[Thompson(2017)]{thompson2017economic}
Neil Thompson.
\newblock \href{https://papers.ssrn.com/sol3/papers.cfm?abstract_id=2899115}{The economic impact of moore's law: Evidence from when it faltered}.
\newblock \emph{Available at SSRN 2899115}, 2017.

\bibitem[Thompson and Spanuth(2021)]{thompson2021decline}
Neil~C Thompson and Svenja Spanuth.
\newblock \href{https://cacm.acm.org/research/the-decline-of-computers-as-a-general-purpose-technology/}{The decline of computers as a general purpose technology}.
\newblock \emph{Communications of the ACM}, 64\penalty0 (3):\penalty0 64--72, 2021.

\bibitem[Thompson et~al.(2020)Thompson, Greenewald, Lee, and Manso]{thompson2020computational}
Neil~C Thompson, Kristjan Greenewald, Keeheon Lee, and Gabriel~F Manso.
\newblock \href{https://arxiv.org/pdf/2007.05558}{The computational limits of deep learning}.
\newblock \emph{arXiv preprint arXiv:2007.05558}, 2020.

\bibitem[Thompson et~al.(2021)Thompson, Greenewald, Lee, and Manso]{thompson2021deep}
Neil~C Thompson, Kristjan Greenewald, Keeheon Lee, and Gabriel~F Manso.
\newblock \href{https://spectrum.ieee.org/deep-learning-computational-cost}{Deep learning's diminishing returns: The cost of improvement is becoming unsustainable}.
\newblock \emph{Ieee Spectrum}, 58\penalty0 (10):\penalty0 50--55, 2021.

\bibitem[Thompson et~al.(2022)Thompson, Ge, and Manso]{thompson2022importance}
Neil~C Thompson, Shuning Ge, and Gabriel~F Manso.
\newblock \href{https://arxiv.org/pdf/2206.14007}{The importance of (exponentially more) computing power}.
\newblock \emph{arXiv preprint arXiv:2206.14007}, 2022.

\bibitem[Vaswani(2017)]{vaswani2017attention}
A~Vaswani.
\newblock \href{https://arxiv.org/pdf/1706.03762}{Attention is all you need}.
\newblock \emph{Advances in Neural Information Processing Systems}, 2017.

\bibitem[Vemprala et~al.(2023)Vemprala, Chen, Shukla, Narayanan, and Kapoor]{survey_grid}
Sai Vemprala, Shuhang Chen, Abhinav Shukla, Dinesh Narayanan, and Ashish Kapoor.
\newblock \href{https://arxiv.org/pdf/2310.00887}{Grid: A platform for general robot intelligence development}, 2023.

\bibitem[Villasevil et~al.(2024)Villasevil, Jain, Macha, Yuan, Ankile, Simeonov, Agrawal, and Gupta]{villasevil2024scaling}
Marcel~Torne Villasevil, Arhan Jain, Vidyaaranya Macha, Jiayi Yuan, Lars~Lien Ankile, Anthony Simeonov, Pulkit Agrawal, and Abhishek Gupta.
\newblock \href{https://openreview.net/pdf?id=UPe0Sjspzr}{Scaling Robot-Learning by Crowdsourcing Simulation Environments}.
\newblock In \emph{RSS 2024 Workshop: Data Generation for Robotics}, 2024.

\bibitem[Wei et~al.(2022)Wei, Tay, Bommasani, Raffel, Zoph, Borgeaud, Yogatama, Bosma, Zhou, Metzler, et~al.]{wei2022emergent}
Jason Wei, Yi~Tay, Rishi Bommasani, Colin Raffel, Barret Zoph, Sebastian Borgeaud, Dani Yogatama, Maarten Bosma, Denny Zhou, Donald Metzler, et~al.
\newblock \href{https://arxiv.org/pdf/2206.07682}{Emergent abilities of large language models}.
\newblock \emph{arXiv preprint arXiv:2206.07682}, 2022.

\bibitem[Wen et~al.(2023)Wen, Lin, So, Chen, Dou, Gao, and Abbeel]{wen2023any}
Chuan Wen, Xingyu Lin, John So, Kai Chen, Qi~Dou, Yang Gao, and Pieter Abbeel.
\newblock \href{https://arxiv.org/pdf/2401.00025}{Any-point trajectory modeling for policy learning}.
\newblock \emph{arXiv preprint arXiv:2401.00025}, 2023.

\bibitem[Wen et~al.(2024)Wen, Zhu, Li, Zhu, Wu, Xu, Liu, Cheng, Shen, Peng, Feng, and Tang]{wen2024tinyvlafastdataefficientvisionlanguageaction}
Junjie Wen, Yichen Zhu, Jinming Li, Minjie Zhu, Kun Wu, Zhiyuan Xu, Ning Liu, Ran Cheng, Chaomin Shen, Yaxin Peng, Feifei Feng, and Jian Tang.
\newblock \href{https://arxiv.org/pdf/2409.12514}{TinyVLA: Towards Fast, Data-Efficient Vision-Language-Action Models for Robotic Manipulation}, 2024.

\bibitem[Wu et~al.(2023)Wu, Antonova, Kan, Lepert, Zeng, Song, Bohg, Rusinkiewicz, and Funkhouser]{wu2023tidybot}
Jimmy Wu, Rika Antonova, Adam Kan, Marion Lepert, Andy Zeng, Shuran Song, Jeannette Bohg, Szymon Rusinkiewicz, and Thomas Funkhouser.
\newblock \href{https://arxiv.org/pdf/2305.05658}{Tidybot: Personalized robot assistance with large language models}.
\newblock \emph{Autonomous Robots}, 47\penalty0 (8):\penalty0 1087--1102, 2023.

\bibitem[Xian et~al.(2024)Xian, Qiao, Xu, Wang, Chen, Zheng, Xiong, Wang, Zhang, Ma, Wang, Dou, Kim, Tian, Chen, Qiu, Lin, He, Si, Zhang, Yang, Liu, Li, Yamazaki, Zhang, Ha, Zhang, Liu, Zheng, Fu, Wu, Geng, Chen, Milky, Hu, Finn, Shi, Liu, Komura, Erickson, Held, Li, Fan, Zhu, Matusik, Gutfreund, Song, Rus, Lin, Zhu, Fragkiadaki, and Gan]{genesis}
Zhou Xian, Yiling Qiao, Zhenjia Xu, Tsun-Hsuan Wang, Zhehuan Chen, Juntian Zheng, Ziyan Xiong, Yian Wang, Mingrui Zhang, Pingchuan Ma, Yufei Wang, Zhiyang Dou, Byungchul Kim, Yunsheng Tian, Yipu Chen, Xiaowen Qiu, Chunru Lin, Tairan He, Zilin Si, Yunchu Zhang, Zhanlue Yang, Tiantian Liu, Tianyu Li, Kashu Yamazaki, Hongxin Zhang, Huy Ha, Yu~Zhang, Michael Liu, Shaokun Zheng, Zipeng Fu, Qi~Wu, Yiran Geng, Feng Chen, Milky, Yuanming Hu, Chelsea Finn, Guanya Shi, Lingjie Liu, Taku Komura, Zackory Erickson, David Held, Minchen Li, Linxi~"Jim" Fan, Yuke Zhu, Wojciech Matusik, Dan Gutfreund, Shuran Song, Daniela Rus, Ming Lin, Bo~Zhu, Katerina Fragkiadaki, and Chuang Gan.
\newblock \href{https://genesis-embodied-ai.github.io/}{Genesis: A Generative and Universal Physics Engine for Robotics and Beyond}, 2024.

\bibitem[Xiao et~al.(2023)Xiao, Liu, Wang, Zhou, Qi, Cheng, He, and Jiang]{survey_xiao2023robot}
Xuan Xiao, Jiahang Liu, Zhipeng Wang, Yanmin Zhou, Yong Qi, Qian Cheng, Bin He, and Shuo Jiang.
\newblock \href{https://arxiv.org/pdf/2311.14379}{Robot learning in the era of foundation models: A survey}.
\newblock \emph{arXiv preprint arXiv:2311.14379}, 2023.

\bibitem[Yang et~al.(2023)Yang, Dong, Liu, Li, Wang, Jiang, Tan, Kang, Zhang, Zhou, et~al.]{yang2023octopus}
Jingkang Yang, Yuhao Dong, Shuai Liu, Bo~Li, Ziyue Wang, Chencheng Jiang, Haoran Tan, Jiamu Kang, Yuanhan Zhang, Kaiyang Zhou, et~al.
\newblock \href{https://arxiv.org/pdf/2310.08588}{Octopus: Embodied vision-language programmer from environmental feedback}.
\newblock \emph{arXiv preprint arXiv:2310.08588}, 2023.

\bibitem[Zeng et~al.(2023)Zeng, Gan, Wang, Liu, and Yu]{zeng2023large}
Fanlong Zeng, Wensheng Gan, Yongheng Wang, Ning Liu, and Philip~S Yu.
\newblock \href{https://arxiv.org/pdf/2311.07226}{Large language models for robotics: A survey}.
\newblock \emph{arXiv preprint arXiv:2311.07226}, 2023.

\bibitem[Zhai et~al.(2022)Zhai, Kolesnikov, Houlsby, and Beyer]{zhai2022scaling}
Xiaohua Zhai, Alexander Kolesnikov, Neil Houlsby, and Lucas Beyer.
\newblock \href{https://arxiv.org/pdf/2106.04560}{Scaling vision transformers}.
\newblock In \emph{Proceedings of the IEEE/CVF conference on computer vision and pattern recognition}, pages 12104--12113, 2022.

\bibitem[Zhou et~al.(2023)Zhou, Dean, Srirama, Rajeswaran, Pari, Hatch, Jain, Yu, Abbeel, Pinto, et~al.]{zhou2023train}
Gaoyue Zhou, Victoria Dean, Mohan~Kumar Srirama, Aravind Rajeswaran, Jyothish Pari, Kyle Hatch, Aryan Jain, Tianhe Yu, Pieter Abbeel, Lerrel Pinto, et~al.
\newblock \href{https://arxiv.org/pdf/2306.00942}{Train offline, test misc: A real robot learning benchmark}.
\newblock pages 9197--9203, 2023.

\bibitem[Zsolt(2024)]{GitHub_Awesome-LLM-Robotics}
Kira Zsolt.
\newblock \href{https://github.com/GT-RIPL/Awesome-LLM-Robotics}{Awesome-LLM-Robotics}, 2024.

\end{thebibliography}

\newpage
\onecolumn
\section{Appendix / supplemental material}

\subsection{Scaling Laws results}
\begin{figure}[h!]
    \centering
    \includegraphics[width=0.4\textwidth]{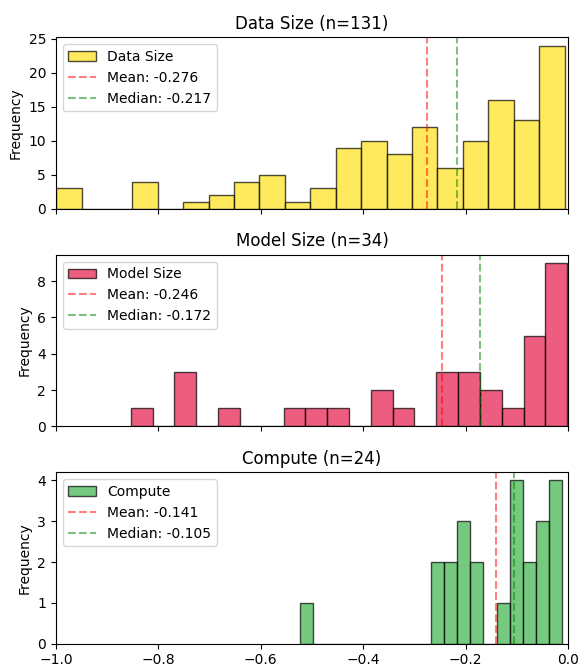}
    \caption{Histograms of \(\alpha\), \(\beta\) and \(\gamma\) values, showing the distributions for data size, model size and compute.}
    \label{fig:Histogram_all_separat}
\end{figure}
\clearpage

\begin{longtable}{lccccc}
    \caption{Data Size Scaling Laws (sorted by \(\alpha\))} \label{tab:Data} \\
    \toprule
    \textbf{Paper} & \textbf{$R^2$} & \textbf{\(\alpha\)} & \textbf{\(A\)} & \textbf{\(E\)} \\
    \midrule
    \endfirsthead

    \multicolumn{5}{c}%
    {{\bfseries Table \thetable\ continued from previous page}} \\
    \toprule
    \textbf{Paper} & \textbf{$R^2$} & \textbf{\(\alpha\)} & \textbf{\(A\)} & \textbf{\(E\)} \\
    \midrule
    \endhead

    \midrule
    \multicolumn{5}{c}{    } \\ 
    \endfoot

    \citep{li2022pre} & 0.74 & -1.00 & 3560.75 & -0.09 \\ 
    \citep{wen2023any} & 0.93 & -1.00 & 104.40 & 54.75 \\ 
    \citep{li2022pre} & 0.96 & -0.98 & 1541.01 & -0.06 \\ 
    \citep{shridhar2022cliport} & 0.90 & -0.84 & 141.94 & -0.09 \\ 
    \citep{radosavovic2023real} & 0.89 & -0.82 & 1269.03 & -14.40 \\ 
    \citep{tang2024embodimentagnosticactionplanningobjectpart} & 1.00 & -0.82 & 111.91 & 26.91 \\ 
    \citep{nair2022r3m} & 1.00 & -0.81 & 219.64 & 37.67 \\ 
    \citep{shridhar2022cliport} & 0.97 & -0.74 & 45.35 & -0.09 \\ 
    \citep{radosavovic2023robot} & 0.82 & -0.67 & 1851.60 & -14.10 \\ 
    \citep{etukuru2024robot} & 0.56 & -0.66 & 4169.55 & -14.40 \\ 
    \citep{stone2023open} & 0.84 & -0.64 & 5137.95 & -1.80 \\ 
    \citep{brohan2022rt} & 0.78 & -0.64 & 68943.94 & -15.30 \\ 
    \citep{li2022pre} & 0.79 & -0.62 & 2597.95 & -2.07 \\ 
    \citep{tang2024embodimentagnosticactionplanningobjectpart} & 0.92 & -0.61 & 34.54 & 46.80 \\ 
    \citep{nair2022r3m} & 1.00 & -0.58 & 177.46 & 25.00 \\ 
    \citep{jiang2023vima} & 0.95 & -0.57 & 3969.39 & 18.63 \\ 
    \citep{jiang2023vima} & 0.96 & -0.57 & 4196.42 & 17.91 \\ 
    \citep{shridhar2022cliport} & 0.99 & -0.56 & 49.30 & -0.38 \\ 
    \citep{wen2023any} & 1.00 & -0.56 & 91.86 & -1.33 \\ 
    \citep{shridhar2022cliport} & 0.98 & -0.52 & 88.16 & -1.08 \\ 
    \citep{tang2024embodimentagnosticactionplanningobjectpart} & 0.68 & -0.49 & 26.06 & -2.17 \\ 
    \citep{nair2022r3m} & 1.00 & -0.46 & 46.43 & 50.40 \\ 
    \citep{shridhar2022cliport} & 0.73 & -0.46 & 19.11 & 0.24 \\ 
    \citep{shridhar2022cliport} & 0.92 & -0.45 & 65.24 & -0.61 \\ 
    \citep{wen2023any} & 1.00 & -0.45 & 140.08 & -14.21 \\ 
    \citep{etukuru2024robot} & 0.88 & -0.44 & 1417.67 & -28.80 \\ 
    \citep{jiang2023vima} & 0.93 & -0.44 & 1922.43 & 12.83 \\ 
    \citep{etukuru2024robot} & 0.72 & -0.44 & 1320.05 & -21.60 \\ 
    \citep{stone2023open} & 0.85 & -0.43 & 5477.52 & -18.90 \\ 
    \citep{rana2024affordancecentricpolicylearningsample} & 0.97 & -0.43 & 201.41 & -18.00 \\ 
    \citep{shridhar2022cliport} & 0.83 & -0.42 & 30.49 & 12.78 \\ 
    \citep{shridhar2022cliport} & 0.97 & -0.40 & 72.29 & 0.45 \\ 
    \citep{fu2024mobile} & 0.97 & -0.40 & 455.12 & -45.00 \\ 
    \citep{shridhar2022cliport} & 1.00 & -0.38 & 48.59 & -1.62 \\ 
    \citep{zhou2023train} & 0.99 & -0.38 & 384.18 & -7.86 \\ 
    \citep{shridhar2022cliport} & 0.89 & -0.38 & 60.61 & -2.88 \\ 
    \citep{jiang2023vima} & 0.95 & -0.37 & 656.99 & 45.36 \\ 
    \citep{shridhar2022cliport} & 1.00 & -0.37 & 93.47 & -3.42 \\ 
    \citep{gao2024efficientdatacollectionrobotic} & 1.00 & -0.37 & 93.35 & 66.60 \\ 
    \citep{gao2024efficientdatacollectionrobotic} & 0.93 & -0.36 & 86.22 & 70.20 \\ 
    \citep{bu2024synergisticgeneralizedefficientdualsystem} & 1.00 & -0.36 & 54.10 & -3.56 \\ 
    \citep{shridhar2022cliport} & 0.99 & -0.35 & 47.39 & 6.75 \\ 
    \citep{nair2022r3m} & 1.00 & -0.35 & 89.77 & 27.82 \\ 
    \citep{kuang2024ramretrievalbasedaffordancetransfer} & 0.96 & -0.35 & 840.46 & 44.35 \\ 
    \citep{shridhar2022cliport} & 0.96 & -0.34 & 29.75 & 0.66 \\ 
    \citep{jiang2023vima} & 0.93 & -0.33 & 458.86 & 52.92 \\ 
    \citep{shridhar2022cliport} & 0.97 & -0.32 & 24.47 & 25.65 \\ 
    \citep{jiang2023vima} & 0.82 & -0.31 & 346.29 & 59.76 \\ 
    \citep{etukuru2024robot} & 0.96 & -0.31 & 612.81 & -34.20 \\ 
    \citep{shridhar2022cliport} & 0.92 & -0.31 & 17.40 & 17.73 \\ 
    \citep{shridhar2022cliport} & 0.99 & -0.29 & 48.34 & -2.79 \\ 
    \citep{shridhar2022cliport} & 1.00 & -0.29 & 47.21 & 41.18 \\ 
    \citep{nair2022r3m} & 0.94 & -0.29 & 108.54 & -19.80 \\ 
    \citep{radosavovic2023robot} & 1.00 & -0.29 & 81.01 & 63.34 \\ 
    \citep{shridhar2022cliport} & 1.00 & -0.28 & 53.77 & 20.16 \\ 
    \citep{nair2022r3m} & 1.00 & -0.28 & 139.44 & -27.00 \\ 
    \citep{nair2022r3m} & 0.97 & -0.28 & 94.36 & -18.00 \\ 
    \citep{etukuru2024robot} & 0.53 & -0.27 & 340.44 & -21.60 \\ 
    \citep{gao2024efficientdatacollectionrobotic} & 1.00 & -0.27 & 68.66 & 64.83 \\ 
    \citep{li2024llarasuperchargingrobotlearning} & 0.97 & -0.27 & 443.58 & -7.21 \\ 
    \citep{jiang2023vima} & 0.95 & -0.26 & 293.92 & 49.37 \\ 
    \citep{shridhar2022cliport} & 0.94 & -0.26 & 75.00 & 0.09 \\ 
    \citep{shridhar2022cliport} & 0.95 & -0.25 & 28.46 & 13.32 \\ 
    \citep{shridhar2022cliport} & 0.98 & -0.23 & 57.60 & -5.40 \\ 
    \citep{nair2022r3m} & 1.00 & -0.22 & 139.22 & -23.40 \\ 
    \citep{shridhar2022cliport} & 0.98 & -0.22 & 41.27 & 20.34 \\ 
    \citep{radosavovic2023robot} & 0.88 & -0.22 & 459.37 & -40.50 \\ 
    \citep{radosavovic2023robot} & 0.80 & -0.21 & 315.23 & -37.80 \\ 
    \citep{li2024llarasuperchargingrobotlearning} & 0.85 & -0.20 & 227.99 & -10.03 \\ 
    \citep{li2024llarasuperchargingrobotlearning} & 0.85 & -0.20 & 227.99 & -10.03 \\ 
    \citep{nair2022r3m} & 0.95 & -0.19 & 108.73 & -27.90 \\ 
    \citep{reed2022generalist} & 0.95 & -0.18 & 49.72 & 29.36 \\ 
    \citep{li2024llarasuperchargingrobotlearning} & 0.86 & -0.18 & 442.45 & -23.19 \\ 
    \citep{shridhar2022cliport} & 0.87 & -0.17 & 54.38 & -8.73 \\ 
    \citep{shridhar2022cliport} & 0.98 & -0.17 & 80.30 & -0.60 \\ 
    \citep{shridhar2022cliport} & 0.77 & -0.16 & 21.56 & 2.52 \\ 
    \citep{wen2023any} & 0.97 & -0.16 & 98.51 & -28.80 \\ 
    \citep{li2024llarasuperchargingrobotlearning} & 0.73 & -0.16 & 195.00 & -12.81 \\ 
    \citep{wen2023any} & 0.95 & -0.15 & 87.50 & -28.80 \\ 
    \citep{shridhar2022cliport} & 0.51 & -0.15 & 17.02 & 36.18 \\ 
    \citep{brohan2022rt} & 0.80 & -0.15 & 430.91 & -36.90 \\ 
    \citep{shridhar2022cliport} & 0.78 & -0.14 & 90.36 & -15.84 \\ 
    \citep{etukuru2024robot} & 0.92 & -0.14 & 209.13 & -34.20 \\ 
    \citep{jiang2023vima} & 0.85 & -0.14 & 57.11 & 79.20 \\ 
    \citep{shridhar2022cliport} & 0.96 & -0.14 & 93.96 & -12.74 \\ 
    \citep{jiang2023vima} & 0.93 & -0.14 & 53.46 & 79.02 \\ 
    \citep{bu2024synergisticgeneralizedefficientdualsystem} & 0.99 & -0.13 & 186.30 & -47.97 \\ 
    \citep{jiang2023vima} & 0.91 & -0.13 & 236.12 & 7.00 \\ 
    \citep{nair2022r3m} & 0.94 & -0.12 & 97.70 & -30.60 \\ 
    \citep{gao2024efficientdatacollectionrobotic} & 1.00 & -0.12 & 221.95 & -45.72 \\ 
    \citep{li2024llarasuperchargingrobotlearning} & 0.89 & -0.12 & 198.21 & -17.50 \\ 
    \citep{bu2024synergisticgeneralizedefficientdualsystem} & 0.95 & -0.12 & 150.23 & -40.28 \\ 
    \citep{shridhar2022cliport} & 0.71 & -0.11 & 57.26 & -12.42 \\ 
    \citep{li2024llarasuperchargingrobotlearning} & 0.96 & -0.11 & 293.84 & -37.73 \\ 
    \citep{nair2022r3m} & 0.96 & -0.11 & 139.93 & -46.80 \\ 
    \citep{nair2022r3m} & 0.99 & -0.10 & 158.69 & -54.00 \\ 
    \citep{nair2022r3m} & 0.99 & -0.10 & 160.53 & -54.90 \\ 
    \citep{nair2022r3m} & 0.99 & -0.10 & 162.36 & -55.80 \\ 
    \citep{li2024llarasuperchargingrobotlearning} & 0.78 & -0.10 & 279.68 & -38.77 \\ 
    \citep{shridhar2022cliport} & 0.93 & -0.09 & 126.32 & -30.06 \\ 
    \citep{li2024llarasuperchargingrobotlearning} & 0.91 & -0.09 & 263.50 & -42.92 \\ 
    \citep{li2024llarasuperchargingrobotlearning} & 0.83 & -0.08 & 152.04 & -26.66 \\ 
    \citep{li2024llarasuperchargingrobotlearning} & 0.83 & -0.08 & 152.04 & -26.66 \\ 
    \citep{shridhar2022cliport} & 0.81 & -0.08 & 75.61 & -21.06 \\ 
    \citep{jiang2023vima} & 0.97 & -0.08 & 242.10 & -38.25 \\ 
    \citep{li2024llarasuperchargingrobotlearning} & 0.79 & -0.07 & 148.91 & -28.73 \\ 
    \citep{jiang2023vima} & 0.94 & -0.07 & 233.21 & -42.48 \\ 
    \citep{nair2022r3m} & 0.99 & -0.06 & 113.13 & -41.40 \\ 
    \citep{jiang2023vima} & 0.95 & -0.05 & 212.85 & -53.73 \\ 
    \citep{jiang2023vima} & 0.95 & -0.05 & 211.74 & -54.72 \\ 
    \citep{jiang2023vima} & 0.80 & -0.04 & 209.03 & -59.58 \\ 
    \citep{shridhar2022cliport} & 0.66 & -0.04 & 143.66 & -50.58 \\ 
    \citep{reed2022generalist} & 0.73 & -0.04 & 156.79 & -49.50 \\ 
    \citep{reed2022generalist} & 0.05 & -0.04 & 9.22 & 31.50 \\ 
    \citep{shridhar2022cliport} & 0.59 & -0.03 & 58.60 & -21.42 \\ 
    \citep{bousmalis2023robocat} & 0.19 & -0.03 & 58.46 & -20.70 \\ 
    \citep{bousmalis2023robocat} & 0.91 & -0.03 & 87.05 & -32.40 \\ 
    \citep{shridhar2022cliport} & 0.99 & -0.03 & 139.00 & -52.83 \\ 
    \citep{shridhar2022cliport} & 0.43 & -0.03 & 58.53 & -20.97 \\ 
    \citep{bousmalis2023robocat} & 0.83 & -0.03 & 51.05 & -19.80 \\ 
    \citep{shridhar2022cliport} & 0.82 & -0.02 & 143.14 & -58.14 \\ 
    \citep{shridhar2022cliport} & 0.78 & -0.02 & 132.32 & -54.00 \\ 
    \citep{shridhar2022cliport} & 0.38 & -0.02 & 101.48 & -40.23 \\ 
    \citep{shridhar2022cliport} & 0.70 & -0.02 & 151.84 & -63.90 \\ 
    \citep{shridhar2022cliport} & 0.01 & -0.01 & 12.49 & 61.47 \\ 
    \citep{tang2024embodimentagnosticactionplanningobjectpart} & 0.68 & -0.01 & 43.45 & -18.70 \\ 
    \citep{shridhar2022cliport} & 0.57 & -0.01 & 72.93 & -31.59 \\ 
    \citep{jiang2023vima} & 0.92 & -0.01 & 193.82 & -77.13 \\ 
    \citep{reed2022generalist} & 0.54 & -0.01 & 77.80 & -34.20 \\ 
    \citep{shridhar2022cliport} & 0.10 & -0.01 & 130.62 & -56.43 \\ 
    \citep{shridhar2022cliport} & 0.23 & -0.01 & 127.58 & -56.70 \\ 
\end{longtable}

\begin{longtable}{lccccc}
    \caption{Model Size Scaling Laws (sorted by \(\beta\))} \label{tab:Model} \\
    \toprule
    \textbf{Paper} & \textbf{$R^2$} & \textbf{\(\beta\)} & \textbf{\(B\)} & \textbf{\(E\)} \\
    \midrule
    \endfirsthead

    \multicolumn{5}{c}%
    {{\bfseries Table \thetable\ continued from previous page}} \\
    \toprule
    \textbf{Paper} & \textbf{$R^2$} & \textbf{\(\beta\)} & \textbf{\(B\)} & \textbf{\(E\)} \\
    \midrule
    \endhead

    \midrule
    \multicolumn{5}{c}{    } \\ 
    \endfoot

    \citep{staroverov2023fine} & 0.97 & -0.85 & 71973454.54 & 9.00 \\ 
    \citep{jiang2023vima} & 0.95 & -0.75 & 19749876.84 & 17.10 \\ 
    \citep{jiang2023vima} & 0.90 & -0.75 & 14716190.30 & 18.00 \\ 
    \citep{chen2023open} & 0.74 & -0.74 & 1831083835.62 & 2.94 \\ 
    \citep{jiang2023vima} & 0.97 & -0.68 & 5822660.53 & 14.42 \\ 
    \citep{stone2023open} & 0.95 & -0.53 & 718180.08 & -4.50 \\ 
    \citep{lynch2020language} & 1.00 & -0.47 & 367339.78 & -7.43 \\ 
    \citep{radosavovic2023real} & 1.00 & -0.44 & 412998.38 & -13.50 \\ 
    \citep{staroverov2023fine} & 1.00 & -0.38 & 12279.99 & 0.43 \\ 
    \citep{stone2023open} & 1.00 & -0.38 & 51670.03 & -0.67 \\ 
    \citep{jiang2023vima} & 0.86 & -0.32 & 3975.06 & 47.34 \\ 
    \citep{jiang2023vima} & 0.99 & -0.24 & 1535.49 & 45.34 \\ 
    \citep{jiang2023vima} & 0.42 & -0.24 & 693.76 & 45.36 \\ 
    \citep{chen2023open} & 1.00 & -0.22 & 8164.69 & -5.17 \\ 
    \citep{huang2023instruct2act} & 0.99 & -0.20 & 3026.16 & -14.40 \\ 
    \citep{jiang2023vima} & 0.51 & -0.18 & 146.26 & 16.65 \\ 
    \citep{radosavovic2023robot} & 0.90 & -0.18 & 4170.81 & -37.80 \\ 
    \citep{jiang2023vima} & 0.68 & -0.17 & 418.20 & 37.86 \\ 
    \citep{jiang2023vima} & 0.72 & -0.13 & 326.99 & 36.51 \\ 
    \citep{chen2023open} & 0.85 & -0.11 & 2309.37 & -36.00 \\ 
    \citep{jiang2023vima} & 0.77 & -0.07 & 464.23 & -51.30 \\ 
    \citep{chen2023open} & 0.29 & -0.05 & 200.67 & -18.00 \\ 
    \citep{jiang2023vima} & 0.35 & -0.05 & 31.93 & 74.88 \\ 
    \citep{jiang2023vima} & 0.95 & -0.05 & 229.89 & -37.80 \\ 
    \citep{cheang2024gr2generativevideolanguageactionmodel} & 0.81 & -0.04 & 197.03 & -36.27 \\ 
    \citep{jiang2023vima} & 0.70 & -0.04 & 190.45 & -41.49 \\ 
    \citep{jiang2023vima} & 0.52 & -0.02 & 55.31 & -16.65 \\ 
    \citep{lynch2020language} & 0.21 & -0.02 & 176.60 & -55.80 \\ 
    \citep{wu2023tidybot} & 0.96 & -0.01 & 148.94 & -45.00 \\ 
    \citep{jiang2023vima} & 0.41 & -0.01 & 118.23 & -47.61 \\ 
    \citep{jiang2023vima} & 0.34 & -0.01 & 129.16 & -52.11 \\ 
    \citep{jiang2023vima} & 0.22 & -0.01 & 111.09 & -45.99 \\ 
    \citep{wu2023tidybot} & 0.99 & -0.01 & 119.03 & -48.60 \\ 
    \citep{jiang2023vima} & 0.13 & -0.00 & 170.34 & -77.22 \\ 
\end{longtable}

\begin{longtable}{lccccc}
    \caption{Compute Scaling Laws (based on \citet{liu2023tail}), sorted by~\(\gamma\)} \label{tab:Compute} \\
    \toprule
    \textbf{Task - Method} & \textbf{$R^2$} & \textbf{\(\gamma\)} & \textbf{\(F\)} & \textbf{\(E\)} \\
    \midrule
    \endfirsthead 

    \multicolumn{5}{c}%
    {{\bfseries Table \thetable\ continued from previous page}} \\
    \toprule
    \textbf{Paper reference} & \textbf{$R^2$} & \textbf{\(\gamma\)} & & \textbf{\(F\)} & \textbf{\(E\)} \\
    \midrule
    \endhead 

    \midrule
    \multicolumn{5}{c}{}\\ 
    \endfoot

        Kitchen - LORA & 0.85 & -0.52 & 1.39E+11 & 11.31 \\  
        Kitchen - Adapter & 0.89 & -0.25 & 3182986.28 & -24.36 \\  
        Goal - Prefix & 0.94 & -0.25 & 1943383.77 & -27.66 \\  
        Spatial - Lora & 0.84 & -0.23 & 725124.6 & -6.39 \\  
        Kitchen - Prefix & 0.98 & -0.23 & 1500921.68 & -28.59 \\  
        Object - Prefix & 0.87 & -0.22 & 655550.95 & -32.35 \\  
        Kitchen - Roboadapter & 0.83 & -0.21 & 628849.75 & -30.37 \\  
        Living - Lora & 0.94 & -0.19 & 189573.61 & -27.19 \\  
        Object - Lora & 0.93 & -0.18 & 146726.06 & -27.66 \\  
        Goal - Lora & 0.88 & -0.17 & 71074.44 & -21.33 \\  
        Goal - Roboadapter & 0.94 & -0.13 & 16331.03 & -15.49 \\  
        Study - Lora & 0.88 & -0.11 & 9581.04 & -43.5 \\  
        Living - Adapter & 0.83 & -0.1 & 8033.2 & -52.5 \\  
        Goal - Adapter & 0.69 & -0.09 & 4388.98 & -34.49 \\  
        Spatial - Prefix & 0.92 & -0.09 & 5269.53 & -53.43 \\  
        Spatial - Adapter & 0.8 & -0.07 & 2035.6 & -39.94 \\  
        Living - Prefix & 0.86 & -0.07 & 2751.37 & -62.94 \\  
        Spatial - Roboadapter & 0.82 & -0.06 & 1678.89 & -59.53 \\  
        Living - Roboadapter & 0.8 & -0.06 & 1524.99 & -60.07 \\  
        Study - Prefix & 0.92 & -0.05 & 1052.19 & -64.68 \\  
        Object - Adapter & 0.36 & -0.03 & 492.04 & -60 \\  
        Study - Roboadapter & 0.83 & -0.03 & 561.34 & -72.19 \\  
        Object - Roboadapter & 0.36 & -0.02 & 412.99 & -65.86 \\  
        Study - Adapter & 0.35 & -0.01 & 266.61 & -74.53 \\  
\end{longtable}


\end{document}